\documentclass{article} 
\usepackage{iclr2019_conference,times}

\usepackage{amsmath,amsfonts,bm}

\def\eqref#1{equation~\ref{#1}}

\def\1{\bm{1}}

\DeclareMathAlphabet{\mathsfit}{\encodingdefault}{\sfdefault}{m}{sl}
\SetMathAlphabet{\mathsfit}{bold}{\encodingdefault}{\sfdefault}{bx}{n}

\newcommand{\R}{\mathbb{R}}

\DeclareMathOperator*{\argmax}{arg\,max}

\usepackage{hyperref}
\usepackage{url}
\usepackage{graphicx}
\usepackage{subcaption}

\iclrfinalcopy

\title{Global-to-local Memory Pointer Networks for Task-Oriented Dialogue}

\author{Chien-Sheng Wu$^\dag$\thanks{All work was done while the first author was an intern at Salesforce Research.} , Richard Socher$^\ddag$ \& Caiming Xiong$^\ddag$ \\
$^\ddag$Salesforce Research \\
\texttt{\{rsocher,cxiong\}@salesforce.com} \\
$^\dag$The Hong Kong University of Science and Technology \\
\texttt{jason.wu@connect.ust.hk} \\
}

\begin{document}

\maketitle

\begin{abstract}
End-to-end task-oriented dialogue is challenging since knowledge bases are usually large, dynamic and hard to incorporate into a learning framework. We propose the global-to-local memory pointer (GLMP) networks to address this issue. In our model, a global memory encoder and a local memory decoder are proposed to share external knowledge. The encoder encodes dialogue history, modifies global contextual representation, and generates a global memory pointer. The decoder first generates a sketch response with unfilled slots. Next, it passes the global memory pointer to filter the external knowledge for relevant information, then instantiates the slots via the local memory pointers. We empirically show that our model can improve copy accuracy and mitigate the common out-of-vocabulary problem. As a result, GLMP is able to improve over the previous state-of-the-art models in both simulated bAbI Dialogue dataset and human-human Stanford Multi-domain Dialogue dataset on automatic and human evaluation.
\end{abstract}

\section{Introduction}
Task-oriented dialogue systems aim to achieve specific user goals such as restaurant reservation or navigation inquiry within a limited dialogue turns via natural language. Traditional pipeline solutions are composed of natural language understanding, dialogue management and natural language generation~\citep{young2013pomdp,wen2016network}, where each module is designed separately and expensively. In order to reduce human effort and scale up between domains, end-to-end dialogue systems, which input plain text and directly output system responses, have shown promising results based on recurrent neural networks ~\citep{zhao2017generative,lei2018sequicity} and memory networks~\citep{sukhbaatar2015end}. These approaches have the advantages that the dialogue states are latent without hand-crafted labels and eliminate the needs to model the dependencies between modules and interpret knowledge bases (KB) manually.

However, despite the improvement by modeling KB with memory network~\citep{bordes2016learning,mem2seq}, end-to-end systems usually suffer from effectively incorporating external KB into the system response generation. The main reason is that a large, dynamic KB is equal to a noisy input and hard to encode and decode, which makes the generation unstable. Different from chit-chat scenario, this problem is especially harmful in task-oriented one, since the information in KB is usually the expected entities in the response. For example, in Table~\ref{TB:EXAMPLE} the driver will expect to get the correct address to the gas station other than a random place such as a hospital. Therefore, pointer networks~\citep{vinyals2015pointer} or copy mechanism~\citep{guEtAl2016} is crucial to successfully generate system responses because directly copying essential words from the input source to the output not only reduces the generation difficulty, but it is also more like a human behavior. For example, in Table~\ref{TB:EXAMPLE}, when human want to reply others the \textit{Valero}'s address, they will need to ``copy'' the information from the table to their response as well. 

Therefore, in the paper, we propose the global-to-local memory pointer (GLMP) networks, which is composed of a global memory encoder, a local memory decoder, and a shared external knowledge. Unlike existing approaches with copy ability~\citep{gulcehreEtAl2016,guEtAl2016,eric-manning:2017:EACLshort,mem2seq}, which the only information passed to decoder is the encoder hidden states, our model shares the external knowledge and leverages the encoder and the external knowledge to learn a global memory pointer and global contextual representation. Global memory pointer modifies the external knowledge by softly filtering words that are not necessary for copying. Afterward, instead of generating system responses directly, the local memory decoder first uses a sketch RNN to obtain sketch responses without slot values but sketch tags, which can be considered as learning a latent dialogue management to generate dialogue action template. Then the decoder generates local memory pointers to copy words from external knowledge and instantiate sketch tags. 

We empirically show that GLMP can achieve superior performance using the combination of global and local memory pointers. In simulated out-of-vocabulary (OOV) tasks in the bAbI dialogue dataset~\citep{bordes2016learning}, GLMP achieves 92.0\% per-response accuracy and surpasses existing end-to-end approaches by 7.5\% in full dialogue. In the human-human dialogue dataset~\citep{ericKVR2017}, GLMP is able to surpass the previous state of the art on both automatic and human evaluation, which further confirms the effectiveness of our double pointers usage.

\begin{table}[t]
\setlength{\abovecaptionskip}{-1pt} 
\caption{An in-car assistant example on the navigation domain. The left part is the KB information and the right part is the conversation between a driver and our system.}
\label{TB:EXAMPLE}
\begin{center}
\resizebox{\linewidth}{!}{
\begin{tabular}{|l|l|l|l|l|ll|l}
\cline{1-5} \cline{7-8}
\textbf{Point of interest (poi)} & \textbf{Distance} & \textbf{Traffic} & \textbf{Poi type} & \textbf{Address} &  & Driver & I need gas \\ \cline{1-5} \cline{7-8} 
Toms house & 3 miles & heavy & friend's house & 580 Van Ness Ave &  & System & GLMP: There is a gas station locally Valero is 4 miles away \\ \cline{1-5}
Coupa & 2 miles & moderate & coffee or tea place & 394 Van Ness Ave &  &  & Gold: Valero is 4 miles away \\ \cline{1-5} \cline{7-8} 
Panda express & 2 miles & no & Chinese restaurant & 842 Arrowhead Way &  & Driver & What is the address ? \\ \cline{1-5} \cline{7-8} 
Stanford express care & 5 miles & no & hospital & 214 El Camino Real &  & System & GLMP: Valero is located at 200 Alester Ave \\ \cline{1-5}
Valero & 4 miles & heavy & gas station & 200 Alester Ave &  &  & Gold: Valero is at 200 Alester Ave \\ \cline{1-5} \cline{7-8} 
Starbucks & 1 miles & heavy & coffee or tea place & 792 Bedoin St &  & Driver & Thank you! \\ \cline{1-5} \cline{7-8} 
\end{tabular}
}
\end{center}
\vspace{-1.0\baselineskip}
\end{table}

\section{GLMP Model}
Our model~\footnote{https://github.com/jasonwu0731/GLMP} is composed of three parts: global memory encoder, external knowledge, and local memory decoder, as shown in Figure~\ref{FIG:MODEL1}(a). The dialogue history $X = (x_1,\dots,x_n)$ and the KB information $B = (b_1,\dots,b_l)$ are the input, and the system response $Y=(y_1,\dots,y_m)$ is the expected output, where $n, l, m$ are the corresponding lengths. First, the global memory encoder uses a context RNN to encode dialogue history and writes its hidden states into the external knowledge. Then the last hidden state is used to read the external knowledge and generate the global memory pointer at the same time. On the other hand, during the decoding stage, the local memory decoder first generates sketch responses by a sketch RNN. Then the global memory pointer and the sketch RNN hidden state are passed to the external knowledge as a filter and a query. The local memory pointer returns from the external knowledge can copy text from the external knowledge to replace the sketch tags and obtain the final system response.

\subsection{External Knowledge}
\begin{figure}[t]
\begin{subfigure}{0.65\textwidth}
  \centering
  \includegraphics[width=0.95\linewidth]{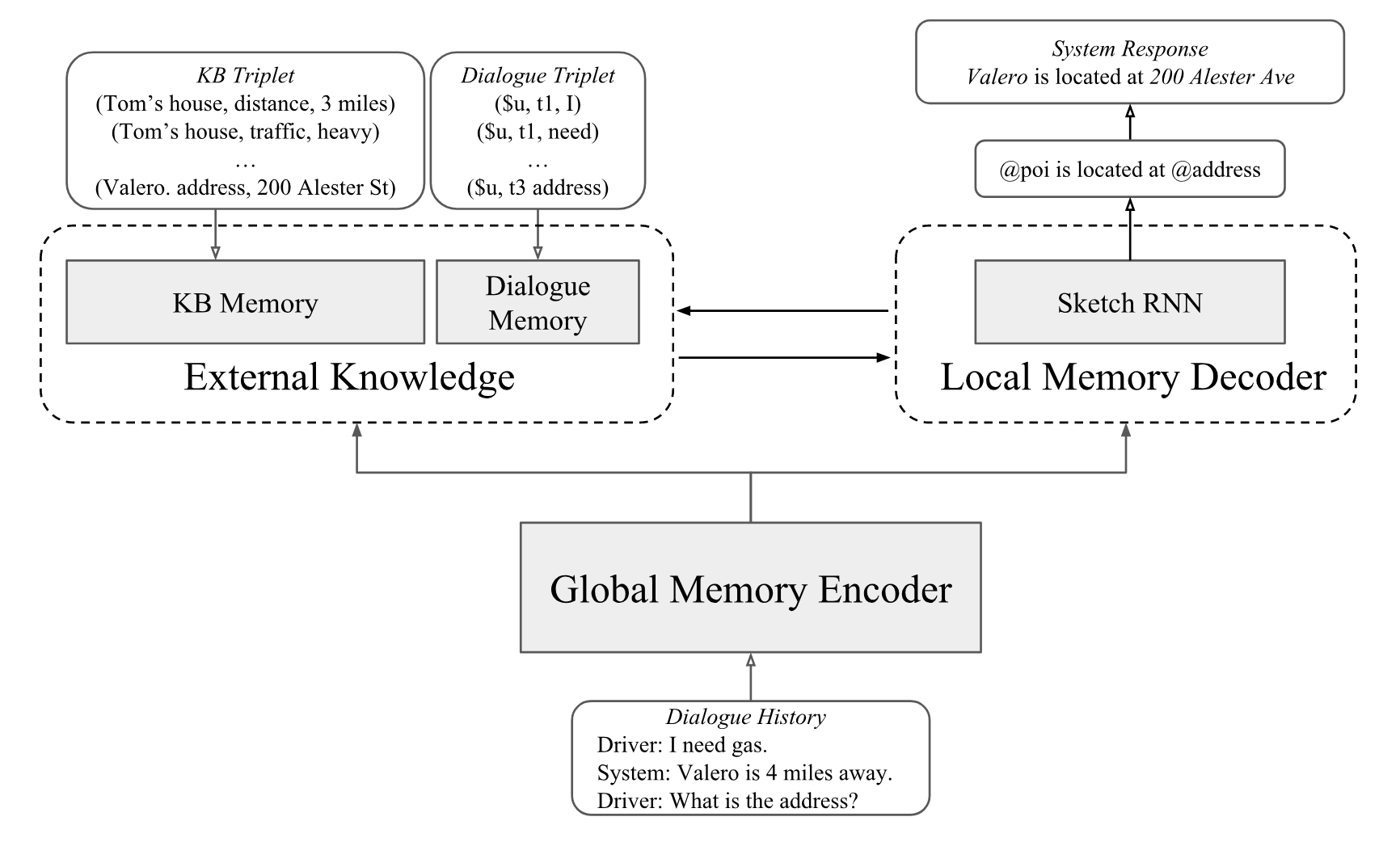}
  \caption{Block diagram}
\end{subfigure}%
\begin{subfigure}{0.35\textwidth}
  \centering
  \includegraphics[width=\linewidth]{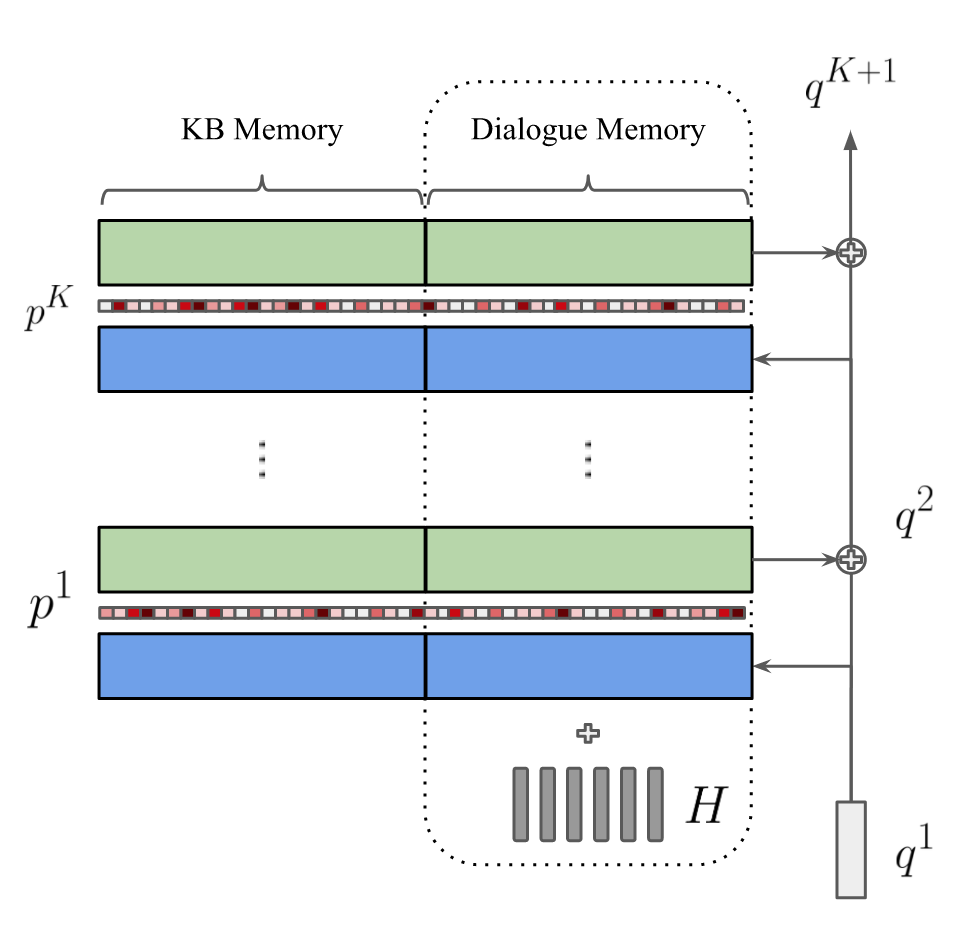}
  \caption{External knowledge}
\end{subfigure}%
\caption{The proposed (a) global-to-local memory pointer networks for task-oriented dialogue systems and the (b) external knowledge architecture. }
\label{FIG:MODEL1}
\end{figure}

Our external knowledge contains the global contextual representation that is shared with the encoder and the decoder. To incorporate external knowledge into a learning framework, end-to-end memory networks (MN) are used to store word-level information for both structural KB (KB memory) and temporal-dependent dialogue history (dialogue memory), as shown in Figure~\ref{FIG:MODEL1}(b). In addition, the MN is well-known for its multiple hop reasoning ability ~\citep{sukhbaatar2015end}, which is appealing to strengthen copy mechanism. 

\textbf{Global contextual representation.} In the KB memory module, each element $b_i \in B$ is represented in the triplet format as (Subject, Relation, Object) structure, which is a common format used to represent KB nodes~\citep{millerEtAl2016,ericKVR2017}. For example, the KB in the Table~\ref{TB:EXAMPLE} will be denoted as $\{$(\textit{Tom's house, distance, 3 miles), ..., (Starbucks, address, 792 Bedoin St)}$\}$. On the other hand, the dialogue context $X$ is stored in the dialogue memory module, where the speaker and temporal encoding are included as in ~\cite{bordes2016learning} like a triplet format. For instance, the first utterance from the driver in the Table~\ref{TB:EXAMPLE} will be denoted as $\{$(\textit{\$user, turn1, I), (\$user, turn1, need), (\$user, turn1, gas)}$\}$. For the two memory modules, a bag-of-word representation is used as the memory embeddings. During the inference time, we copy the object word once a memory position is pointed to, for example, \textit{3 miles} will be copied if the triplet \textit{(Toms house, distance, 3 miles)} is selected. We denote $Object(.)$ function as getting the object word from a triplet.

\textbf{Knowledge read and write.} Our external knowledge is composed of a set of trainable embedding matrices $C = (C^1,\dots,C^{K+1})$, where $C^k \in \R^{|V|\times d_{emb}}$, $K$ is the maximum memory hop in the MN, $|V|$ is the vocabulary size and $d_{emb}$ is the embedding dimension. We denote memory in the external knowledge as $M = [B; X] = (m_1,\dots,m_{n+l})$, where $m_i$ is one of the triplet components mentioned. To read the memory, the external knowledge needs a initial query vector $q^1$. Moreover, it can loop over $K$ hops and computes the attention weights at each hop $k$ using 
\begin{equation}
  p^k_i = \text{Softmax}((q^k)^T c^k_i),
  \label{attn_eq}
\end{equation}
where $c^k_i = B(C^k(m_i)) \in \R^{d_{emb}}$ is the embedding in $i^{th}$ memory position using the embedding matrix $C^k$, $q^k$ is the query vector for hop $k$, and $B(.)$ is the bag-of-word function. Note that $p^k \in \R^{n+l}$ is a soft memory attention that decides the memory relevance with respect to the query vector. Then, the model reads out the memory $o^k$ by the weighted sum over $c^{k+1}$ and update the query vector $q^{k+1}$. Formally,
\begin{equation}
    o^k = \sum_i p^k_i c^{k+1}_i, \quad q^{k+1} = q^{k} + o^{k}.
\end{equation}

\subsection{Global Memory Encoder}
\begin{figure}[t]
\begin{subfigure}{0.35\textwidth}
  \centering
  \includegraphics[width=\linewidth]{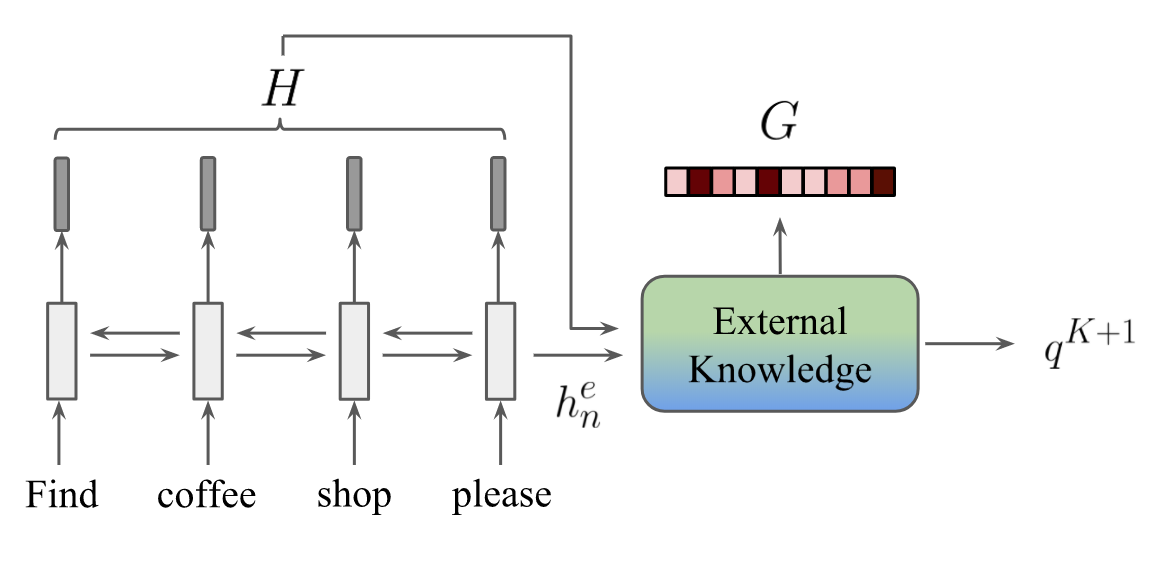}
  \caption{Global memory encoder}
\end{subfigure}%
\begin{subfigure}{0.65\textwidth}
  \centering
  \includegraphics[width=\linewidth]{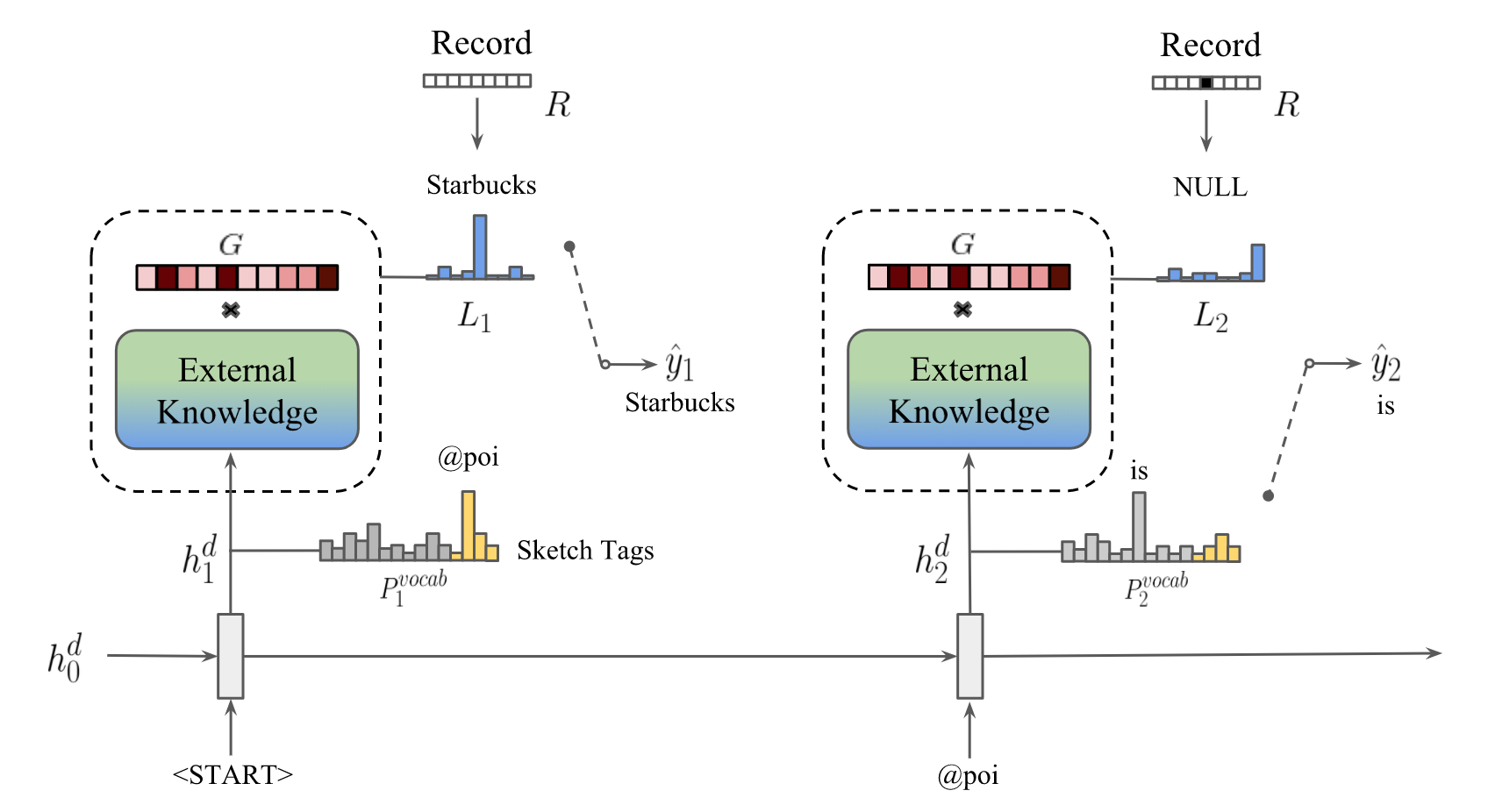}
  \caption{Local memory decoder}
\end{subfigure}%
\caption{The proposed (a) global memory encoder and the (b) local memory decoder architecture. }
\label{FIG:MODEL2}
\end{figure}

In Figure~\ref{FIG:MODEL2}(a), a context RNN is used to model the sequential dependency and encode the context $X$. Then the hidden states are written into the external knowledge as shown in Figure~\ref{FIG:MODEL1}(b). Afterward, the last encoder hidden state serves as the query to read the external knowledge and get two outputs, the global memory pointer and the memory readout. Intuitively, since it is hard for MN architectures to model the dependencies between memories~\citep{dqmem8461426}, which is a serious drawback especially in conversational related tasks, writing the hidden states to the external knowledge can provide sequential and contextualized information. With meaningful representation, our pointers can correctly copy out words from external knowledge, and the common OOV challenge can be mitigated. In addition, using the encoded dialogue context as a query can encourage our external knowledge to read out memory information related to the hidden dialogue states or user intention. Moreover, the global memory pointer that learns a global memory distribution is passed to the decoder along with the encoded dialogue history and KB information.

\textbf{Context RNN.}
A bi-directional gated recurrent unit (GRU)~\citep{GRU} is used to encode dialogue history into the hidden states $H = (h_1^e,\dots,h_1^e)$, and the last hidden state $h_n^e$ is used to query the external knowledge as the encoded dialogue history. In addition, the hidden states $H$ are written into the dialogue memory module in the external knowledge by summing up the original memory representation with the corresponding hidden states. In formula,
\begin{equation}
    \begin{array}{c}
        c^k_i = c^k_i + h^e_{m_i}  \quad \textrm{if} \quad m_i \in X \text{ and } \forall k \in [1,K+1], \\
    \end{array}
\end{equation}
\textbf{Global memory pointer.}
Global memory pointer $G = (g_1,\dots,g_{n+l})$ is a vector containing real values between 0 and 1. Unlike conventional attention mechanism that all the weights sum to one, each element in $G$ is an independent probability. We first query the external knowledge using $h^e_n$ until the last hop, and instead of applying the Softmax function as in (\ref{attn_eq}), we perform an inner product followed by the Sigmoid function. The memory distribution we obtained is the global memory pointer $G$, which is passed to the decoder. To further strengthen the global pointing ability, we add an auxiliary loss to train the global memory pointer as a multi-label classification task. We show in the ablation study that adding this additional supervision does improve the performance. Lastly, the memory readout $q^{K+1}$ is used as the encoded KB information.

In the auxiliary task, we define the label $G^{label} = (g^l_1,\dots,g^l_{n+l})$ by checking whether the object words in the memory exists in the expected system response $Y$. Then the global memory pointer is trained using binary cross-entropy loss $Loss_{g}$ between $G$ and $G^{label}$. In formula,
\begin{equation}
    \begin{array}{c}
        g_i = \text{Sigmoid}((q^K)^T c^K_i), \quad 
        {g^l_i} = 
        \begin{cases} 
        1 &\mbox{if } Object(m_i) \in Y \\ 
        0 &\mbox{otherwise} 
        \end{cases}, \\
        Loss_g = -\sum_{i=1}^{n+l} [g^l_i \times \log{g_i}+(1-g^l_i) \times \log{(1-g_i)}]. 
    \end{array}
\end{equation}

\subsection{Local memory decoder}
Given the encoded dialogue history $h^e_n$, the encoded KB information $q^{K+1}$, and the global memory pointer $G$, our local memory decoder first initializes its sketch RNN using the concatenation of $h^e_n$ and $q^{K+1}$, and generates a sketch response that excludes slot values but includes the sketch tags. For example, sketch RNN will generate ``\textit{@poi is @distance away}'', instead of ``\textit{Starbucks is 1 mile away.}'' At each decoding time step, the hidden state of the sketch RNN is used for two purposes: 1) predict the next token in vocabulary, which is the same as standard sequence-to-sequence (S2S) learning; 2) serve as the vector to query the external knowledge. If a sketch tag is generated, the global memory pointer is passed to the external knowledge, and the expected output word will be picked up from the local memory pointer. Otherwise, the output word is the word that generated by the sketch RNN. For example in Figure~\ref{FIG:MODEL2}(b), a @poi tag is generated at the first time step, therefore, the word \textit{Starbucks} is picked up from the local memory pointer as the system output word.

\textbf{Sketch RNN.}
We use a GRU to generate a sketch response $Y^s = (y^s_1,\dots,y^s_m)$ without real slot values. The sketch RNN learns to generate a dynamic dialogue action template based on the encoded dialogue ($h^e_n$) and KB information ($q^{K+1}$). At each decoding time step $t$, the sketch RNN hidden state $h^d_t$ and its output distribution $P^{vocab}_t$ are defined as
\begin{equation}
h^d_t = \text{GRU}(C^1(\hat{y}^s_{t-1}), h^d_{t-1}), \quad
P^{vocab}_t = \text{Softmax}(W h^d_t) 
\end{equation}
We use the standard cross-entropy loss to train the sketch RNN, we define $Loss_v$ as. 
\begin{equation}
Loss_v = \sum_{t=1}^{m} - \log(P^{vocab}_t(y^s_t)).
\end{equation}
We replace the slot values in $Y$ into sketch tags based on the provided entity table. The sketch tags $ST$ are all the possible slot types that start with a special token, for example, \textit{@address} stands for all the addresses and \textit{@distance} stands for all the distance information. 

\textbf{Local memory pointer.} 
Local memory pointer $L = (L_1,\dots,L_m)$ contains a sequence of pointers. At each time step $t$, the global memory pointer $G$ first modify the global contextual representation using its attention weights,  
\begin{equation}
    \begin{array}{c}
        c^k_i = c^k_i \times g_i, \quad \forall i \in [1,n+l] \text{ and } \forall k \in [1,K+1],
    \end{array}
\end{equation}
and then the sketch RNN hidden state $h^d_t$ queries the external knowledge. The memory attention in the last hop is the corresponding local memory pointer $L_t$, which is represented as the memory distribution at time step $t$. To train the local memory pointer, a supervision on top of the last hop memory attention in the external knowledge is added. We first define the position label of local memory pointer $L^{label}$ at the decoding time step $t$ as
\begin{equation}
    L^{label}_t = 
    \begin{cases} 
        max(z) &\mbox{if } \exists z \ \text{s.t.} \ y_t = Object(m_z), \\ 
        n+l+1  &\mbox{otherwise.} 
    \end{cases} 
\end{equation}
The position $n$+$l$+1 is a null token in the memory that allows us to calculate loss function even if $y_t$ does not exist in the external knowledge. Then, the loss between $L$ and $L^{label}$ is defined as 
\begin{equation}
Loss_l = \sum_{t=1}^{m} - \log(L_t(L^{label}_t)).
\end{equation}
Furthermore, a record $R \in \R^{n+l}$ is utilized to prevent from copying same entities multiple times. All the elements in $R$ are initialized as 1 in the beginning. During the decoding stage, if a memory position has been pointed to, its corresponding position in $R$ will be masked out. During the inference time, $\hat{y}_t$ is defined as
\begin{equation}
    \hat{y}_t = 
    \begin{cases} 
        \argmax(P^{vocab}_t) &\mbox{if } \argmax(P^{vocab}_t) \not\in ST, \\ 
        Object(m_{\argmax(L_t \odot R)})  &\mbox{otherwise,} 
    \end{cases} 
\end{equation}
where $\odot$ is the element-wise multiplication. Lastly, all the parameters are jointly trained by minimizing the weighted-sum of three losses ($\alpha, \beta, \gamma$ are hyper-parameters):
\begin{equation}
Loss = \alpha Loss_g + \beta Loss_v + \gamma Loss_l 
\end{equation}

\section{Experiments}
\subsection{Datasets}
We use two public multi-turn task-oriented dialogue datasets to evaluate our model: the bAbI dialogue ~\citep{bordes2016learning} and Stanford multi-domain dialogue (SMD) ~\citep{ericKVR2017}. The bAbI dialogue includes five simulated tasks in the restaurant domain. Task 1 to 4 are about calling API calls, modifying API calls, recommending options, and providing additional information, respectively. Task 5 is the union of tasks 1-4. There are two test sets for each task: one follows the same distribution as the training set and the other has OOV entity values. On the other hand, SMD is a human-human, multi-domain dialogue dataset. It has three distinct domains: calendar scheduling, weather information retrieval, and point-of-interest navigation. The key difference between these two datasets is, the former has longer dialogue turns but the regular user and system behaviors, the latter has few conversational turns but variant responses, and the KB information is much more complicated. 

\subsection{Training details}
The model is trained end-to-end using Adam optimizer~\citep{KingmaB14}, and learning rate annealing starts from $1e^{-3}$ to $1e^{-4}$. The number of hop $K$ is set to 1,3,6 to compare the performance difference. The weights $\alpha, \beta, \gamma$ summing up the three losses are set to 1. All the embeddings are initialized randomly, and a simple greedy strategy is used without beam-search during the decoding stage. The hyper-parameters such as hidden size and dropout rate are tuned with grid-search over the development set (per-response accuracy for bAbI Dialogue and BLEU score for the SMD).  In addition, to increase model generalization and simulate OOV setting, we randomly mask a small number of input source tokens into an unknown token. The model is implemented in PyTorch and the hyper-parameters used for each task and the dataset statistics are reported in the Appendix.

\begin{table}[t]
\setlength{\abovecaptionskip}{-3pt} 
\caption{Per-response accuracy and completion rate (in the parentheses) on bAbI dialogues. GLMP achieves the least out-of-vocabulary performance drop. Baselines are reported from Query Reduction Network~\citep{seo2016query}, End-to-end Memory Network~\citep{bordes2016learning}, Gated Memory Network~\citep{perez2016gated}, Point to Unknown Word~\citep{gulcehreEtAl2016}, and Memory-to-Sequence~\citep{mem2seq}.}
\label{TB:BABI}
\begin{center}
\resizebox{\linewidth}{!}{
\begin{tabular}{r|cccccc|ccc}
\hline
\multicolumn{1}{c|}{\textbf{Task}} & QRN & MN & GMN & S2S+Attn & Ptr-Unk & Mem2Seq & GLMP K1 & \multicolumn{1}{l}{GLMP K3} & \multicolumn{1}{l}{GLMP K6} \\ \hline
T1 & 99.4 (-) & 99.9 (99.6) & 100 (100) & 100 (100) & 100 (100) & 100 (100) & 100 (100) & 100 (100) & 100 (100) \\
T2 & 99.5 (-) & 100 (100) & 100 (100) & 100 (100) & 100 (100) & 100 (100) & 100 (100) & 100 (100) & 100 (100) \\
T3 & 74.8 (-) & 74.9 (2.0) & 74.9 (0) & 74.8 (0) & 85.1 (19.0) & 94.7 (62.1) & \textbf{96.3 (75.6)} & 96.0 (69.4) & 96.0 (68.7) \\
T4 & 57.2 (-) & 59.5 (3.0) & 57.2 (0) & 57.2 (0) & 100 (100) & 100 (100) & 100 (100) & 100 (100) & 100 (100) \\
T5 & \textbf{99.6 (-)} & 96.1 (49.4) & 96.3 (52.5) & 98.4 (87.3) & 99.4 (91.5) & 97.9 (69.6) & 99.2 (88.5) & 99.0 (86.5) & 99.2 (89.7) \\ \hline
T1 oov & 83.1 (-) & 72.3 (0) & 82.4 (0) & 81.7 (0) & 92.5 (54.7) & 94.0 (62.2) & \textbf{100 (100)} & \textbf{100 (100)} & 99.3 (95.9) \\
T2 oov & 78.9 (-) & 78.9 (0) & 78.9 (0) & 78.9 (0) & 83.2 (0) & 86.5 (12.4) & \textbf{100 (100)} & \textbf{100 (100)} & 99.4 (94.6) \\
T3 oov & 75.2 (-) & 74.4 (0) & 75.3 (0) & 75.3 (0) & 82.9 (13.4) & 90.3 (38.7) & 95.5 (65.7) & \textbf{96.7 (72.9)} & 95.9 (67.7) \\
T4 oov & 56.9 (-) & 57.6 (0) & 57.0 (0) & 57.0 (0) & 100 (100) & 100 (100) & 100 (100) & 100 (100) & 100 (100) \\
T5 oov & 67.8 (-) & 65.5 (0) & 66.7 (0) & 65.7 (0) & 73.6 (0) & 84.5 (2.3) & \textbf{92.0 (21.7)} & 91.0 (17.7) & 91.8 (21.4) \\ \hline
\end{tabular}
}
\end{center}
\end{table}

\subsection{Results}
\textbf{bAbI Dialogue. }
In Table~\ref{TB:BABI}, we follow~\citet{bordes2016learning} to compare the performance based on per-response accuracy and task-completion rate. Note that for utterance retrieval methods, such as QRN, MN, and GMN, cannot correctly recommend options (T3) and provide additional information (T4), and a poor generalization ability is observed in OOV setting, which has around 30\% performance difference in Task 5. Although previous generation-based approaches (Ptr-Unk, Mem2Seq) have mitigated the gap by incorporating copy mechanism, the simplest cases such as generating and modifying API calls (T1, T2) still face a 6-17\% OOV performance drop. On the other hand, GLMP achieves a highest 92.0\% task-completion rate in full dialogue task and surpasses other baselines by a big margin especially in the OOV setting. No per-response accuracy loss for T1, T2, T4 using only the single hop, and only decreases 7-9\% in task 5. 

\begin{table}[t]
\setlength{\abovecaptionskip}{-3pt} 
\caption{In SMD dataset, our model achieves highest BLEU score and entity F1 score over baselines, including previous state-of-the-art result from ~\cite{mem2seq}. (Models with * are reported from ~\cite{ericKVR2017}, where the problem is simplified to the canonicalized forms.)}
\label{TB:SMD}
\begin{center}
\resizebox{\linewidth}{!}{
\begin{tabular}{rccccccccc}
\multicolumn{10}{c}{Automatic Evaluation} \\ \hline
\multicolumn{1}{l|}{} & Rule-Based* & \multicolumn{1}{c|}{KVR*} & S2S & S2S + Attn & Ptr-Unk & \multicolumn{1}{c|}{Mem2Seq} & GLMP K1 & GLMP K3 & GLMP K6 \\ \hline
\multicolumn{1}{r|}{BLEU} & 6.6 & \multicolumn{1}{c|}{13.2} & 8.4 & 9.3 & 8.3 & \multicolumn{1}{c|}{12.6} & 13.83 & \textbf{14.79} & 12.37 \\ \hline
\multicolumn{1}{r|}{Entity F1} & 43.8 & \multicolumn{1}{c|}{48.0} & 10.3 & 19.9 & 22.7 & \multicolumn{1}{c|}{33.4} & 57.25 & \textbf{59.97} & 53.54 \\ \hline
\multicolumn{1}{r|}{Schedule F1} & 61.3 & \multicolumn{1}{c|}{62.9} & 9.7 & 23.4 & 26.9 & \multicolumn{1}{c|}{49.3} & 68.74 & \textbf{69.56} & 69.38 \\
\multicolumn{1}{r|}{Weather F1} & 39.5 & \multicolumn{1}{c|}{47.0} & 14.1 & 25.6 & 26.7 & \multicolumn{1}{c|}{32.8} & 60.87 & \textbf{62.58} & 55.89 \\
\multicolumn{1}{r|}{Navigation F1} & 40.4 & \multicolumn{1}{c|}{41.3} & 7.0 & 10.8 & 14.9 & \multicolumn{1}{c|}{20.0} & 48.62 & \textbf{52.98} & 43.08 \\ \hline
\multicolumn{1}{l}{} & \multicolumn{1}{l}{} & \multicolumn{1}{l}{} &  & \multicolumn{1}{l}{} & \multicolumn{1}{l}{} & \multicolumn{1}{l}{} & \multicolumn{1}{l}{} & \multicolumn{1}{l}{} & \multicolumn{1}{l}{} \\
\multicolumn{10}{c}{Human Evaluation} \\ \hline
\multicolumn{1}{l|}{} & \multicolumn{3}{c|}{Mem2Seq} & \multicolumn{3}{c|}{GLMP} & \multicolumn{3}{c}{Human} \\ \hline
\multicolumn{1}{r|}{Appropriate} & \multicolumn{3}{c|}{3.89} & \multicolumn{3}{c|}{4.15} & \multicolumn{3}{c}{4.6} \\
\multicolumn{1}{r|}{Humanlike} & \multicolumn{3}{c|}{3.80} & \multicolumn{3}{c|}{4.02} & \multicolumn{3}{c}{4.54} \\ \hline
\end{tabular}
}
\end{center}
\vspace{-1.0\baselineskip}
\end{table}

\begin{table}[t]
\setlength{\abovecaptionskip}{-2pt} 
\caption{Ablation study using single hop model.}
\label{TB:ABLATION}
\begin{center}
\resizebox{0.75\linewidth}{!}{
\begin{tabular}{r|ccccc|c}
\hline
\multicolumn{1}{l|}{} & \multicolumn{5}{c|}{\begin{tabular}[c]{@{}c@{}}bAbI Dialogue OOV\\ Per-response Accuracy\end{tabular}} & \begin{tabular}[c]{@{}c@{}}SMD\\ Entity F1\end{tabular} \\ \cline{2-7} 
 & T1 & T2 & T3 & T4 & T5 & All \\ \hline
GLMP & 100 (-) & 100 (-) & 95.5 (-) & 100 (-) & 92.0 (-) & 57.25 (-) \\
GLMP w/o H & 90.4 (-9.6) & 85.6 (-14.4) & 95.4 (-0.1) & 100 (-0) & 86.2 (-5.3) & 47.96 (-9.29) \\
GLMP w/o G & 100 (-0) & 91.7 (-8.3) & 95.5 (-0) & 100 (-0) & 92.4 (+0.4) & 45.78 (-11.47) \\ \hline
\end{tabular}
}
\end{center}
\vspace{-1.0\baselineskip}
\end{table}

\textbf{Stanford Multi-domain Dialogue. }
For human-human dialogue scenario, we follow previous dialogue works~\citep{ericKVR2017,zhao2017generative,mem2seq} to evaluate our system on two automatic evaluation metrics, BLEU and entity F1 score~\footnote{BLEU: \texttt{multi-bleu.perl} script; Entity F1: Micro-average over responses.}. As shown in Table~\ref{TB:SMD}, GLMP achieves a highest 14.79 BLEU and 59.97\% entity F1 score, which is a slight improvement in BLEU but a huge gain in entity F1. In fact, for unsupervised evaluation metrics in task-oriented dialogues, we argue that the entity F1 might be a more comprehensive evaluation metric than per-response accuracy or BLEU, as shown in ~\cite{ericKVR2017} that humans are able to choose the right entities but have very diversified responses. Note that the results of rule-based and KVR are not directly comparable because they simplified the task by mapping the expression of entities to a canonical form using named entity recognition and linking~\footnote{For example, they compared in ``\textit{@poi is @poi\_distance away},'' instead of ``\textit{Starbucks is 1\_mile away.''}}.

Moreover, human evaluation of the generated responses is reported. We compare our work with previous state-of-the-art model Mem2Seq~\footnote{Mem2Seq code is released and we achieve similar results stated in the original paper.} and the original dataset responses as well. We randomly select 200 different dialogue scenarios from the test set to evaluate three different responses. Amazon Mechanical Turk is used to evaluate system appropriateness and human-likeness on a scale from 1 to 5. As the results shown in Table~\ref{TB:SMD}, we see that GLMP outperforms Mem2Seq in both measures, which is coherent to previous observation. We also see that human performance on this assessment sets the upper bound on scores, as expected. More details about the human evaluation are reported in the Appendix.

\textbf{Ablation Study. }
The contributions of the global memory pointer $G$ and the memory writing of dialogue history $H$ are shown in Table~\ref{TB:ABLATION}. We compare the results using GLMP with $K = 1$ in bAbI OOV setting and SMD. GLMP without $H$ means that the context RNN in the global memory encoder does not write the hidden states into the external knowledge. As one can observe, our model without $H$ has 5.3\% more loss in the full dialogue task. On the other hand, GLMP without $G$ means that we do not use the global memory pointer to modify the external knowledge, and an 11.47\% entity F1 drop can be observed in SMD dataset. Note that a 0.4\% increase can be observed in task 5, it suggests that the use of global memory pointer may impose too strong prior entity probability. Even if we only report one experiment in the table, this OOV generalization problem can be mitigated by increasing the dropout ratio during training. 

\begin{figure}[t]
\centering
\includegraphics[width=\linewidth]{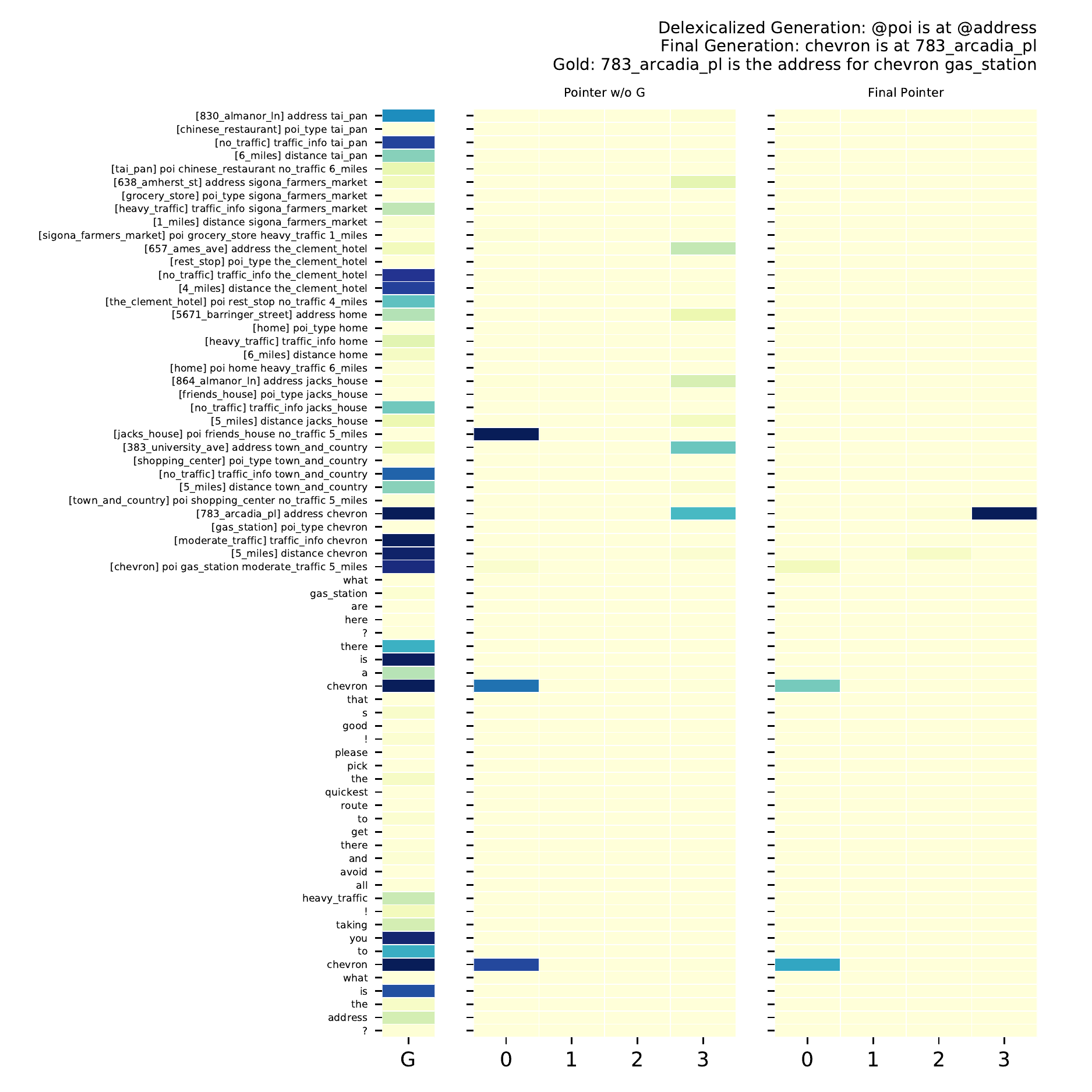} 
\setlength{\abovecaptionskip}{-5pt} 
\setlength{\belowcaptionskip}{-10pt} 
\caption{Memory attention visualization in the SMD navigation domain. Left column is the global memory pointer $G$, middle column is the memory pointer without global weighting, and the right column is the final memory pointer.}
\label{FIG:VIZ}
\end{figure}

\textbf{Visualization and Qualitative Evaluation. }
Analyzing the attention weights has been frequently used to interpret deep learning models. In Figure~\ref{FIG:VIZ}, we show the attention vector in the last hop for each generation time step. Y-axis is the external knowledge that we can copy, including the KB information and the dialogue history. Based on the question ``\textit{what is the address?}'' asked by the driver in the last turn, the gold answer and our generated response are on the top, and the global memory pointer $G$ is shown in the left column. One can observe that in the right column, the final memory pointer successfully copy the entity \textit{chevron} in step 0 and its address \textit{783 Arcadia Pl} in step 3 to fill in the sketch utterance. On the other hand, the memory attention without global weighting is reported in the middle column. One can find that even if the attention weights focus on several point of interests and addresses in step 0 and step 3, the global memory pointer can mitigate the issue as expected. More dialogue visualization and generated results including several negative examples and error analysis are reported in the Appendix.

\section{Related Works}
\textbf{Task-oriented dialogue systems.}
Machine learning based dialogue systems are mainly explored by following two different approaches: modularized and end-to-end. For the modularized systems~\citep{williams2007partially, wen2016network}, a set of modules for natural language understanding~\citep{young2013pomdp, chen2016end}, dialogue state tracking~\citep{lee2016task, zhong2018global}, dialogue management~\citep{su2016line}, and natural language generation~\citep{sharma2016natural} are used. These approaches achieve good stability via combining domain-specific knowledge and slot-filling techniques, but additional human labels are needed. On the other hand, end-to-end approaches have shown promising results recently. Some works view the task as a next utterance retrieval problem, for examples, recurrent entity networks share parameters between RNN~\citep{wu2017dstc6}, query reduction networks modify query between layers~\citep{seo2016query}, and memory networks ~\citep{bordes2016learning, perez2016gated,dqmem8461426} perform multi-hop design to strengthen reasoning ability. In addition, some approaches treat the task as a sequence generation problem. ~\cite{lei2018sequicity} incorporates explicit dialogue states tracking into a delexicalized sequence generation. ~\cite{serban2016building,zhao2017generative} use recurrent neural networks to generate final responses and achieve good results as well. Although it may increase the search space, these approaches can encourage more flexible and diverse system responses by generating utterances token-by-token.

\textbf{Pointer network.}
\cite{vinyals2015pointer} uses attention as a pointer to select a member of the input source as the output. Such copy mechanisms have also been used in other natural language processing tasks, such as question answering~\citep{Dehghani2017, heEtAl2017Long1}, neural machine translation~\citep{gulcehreEtAl2016,guEtAl2016}, language modeling~\citep{merity2016pointer}, and text summarization~\citep{seeliumanning2017Long}. In task-oriented dialogue tasks, \cite{eric-manning:2017:EACLshort} first demonstrated the potential of the copy-augmented Seq2Seq model, which shows that generation-based methods with simple copy strategy can surpass retrieval-based ones. Later, \cite{ericKVR2017} augmented the vocabulary distribution by concatenating KB attention, which at the same time increases the output dimension. Recently, \cite{mem2seq} combines end-to-end memory network into sequence generation, which shows that the multi-hop mechanism in MN can be utilized to improve copy attention. These models outperform utterance retrieval methods by copying relevant entities from the KBs.

\textbf{Others.}
~\cite{zhao2017generative} proposes entity indexing and ~\cite{dqmem8461426} introduces recorded delexicalization to simplify the problem by record entity tables manually. In addition, our approach utilized recurrent structures to query external memory can be viewed as the memory controller in Memory augmented neural networks (MANN)~\citep{graves2014neural, graves2016hybrid}. Similarly, memory encoders have been used in neural machine translation~\citep{wangEtAl2016} and meta-learning applications~\citep{KaiserNRB17}. However, different from other models that use a single matrix representation for reading and writing, GLMP leverages end-to-end memory networks to perform multiple hop attention, which is similar to the stacking self-attention strategy in the Transformer~\citep{vaswani2017attention}.

\section{Conclusion}
In the work, we present an end-to-end trainable model called global-to-local memory pointer networks for task-oriented dialogues. The global memory encoder and the local memory decoder are designed to incorporate the shared external knowledge into the learning framework. We empirically show that the global and the local memory pointer are able to effectively produce system responses even in the out-of-vocabulary scenario, and visualize how global memory pointer helps as well. As a result, our model achieves state-of-the-art results in both the simulated and the human-human dialogue datasets, and holds potential for extending to other tasks such as question answering and text summarization.


\bibliography{iclr2019_conference}

\begin{thebibliography}{35}
\providecommand{\natexlab}[1]{#1}
\providecommand{\url}[1]{\texttt{#1}}
\expandafter\ifx\csname urlstyle\endcsname\relax
  \providecommand{\doi}[1]{doi: #1}\else
  \providecommand{\doi}{doi: \begingroup \urlstyle{rm}\Url}\fi

\bibitem[Bordes \& Weston(2017)Bordes and Weston]{bordes2016learning}
Antoine Bordes and Jason Weston.
\newblock Learning end-to-end goal-oriented dialog.
\newblock \emph{International Conference on Learning Representations},
  abs/1605.07683, 2017.

\bibitem[Chen et~al.(2016)Chen, Hakkani-T{\"u}r, Gao, and Deng]{chen2016end}
Yun-Nung Chen, Dilek Hakkani-T{\"u}r, Jianfeng Gao, and Li~Deng.
\newblock End-to-end memory networks with knowledge carryover for multi-turn
  spoken language understanding.
\newblock 2016.

\bibitem[Chung et~al.(2014)Chung, Gulcehre, Cho, and Bengio]{GRU}
Junyoung Chung, Caglar Gulcehre, Kyunghyun Cho, and Yoshua Bengio.
\newblock Empirical evaluation of gated recurrent neural networks on sequence
  modeling.
\newblock \emph{NIPS Deep Learning and Representation Learning Workshop}, 2014.

\bibitem[Dehghani et~al.(2017)Dehghani, Rothe, Alfonseca, and
  Fleury]{Dehghani2017}
Mostafa Dehghani, Sascha Rothe, Enrique Alfonseca, and Pascal Fleury.
\newblock Learning to attend, copy, and generate for session-based query
  suggestion.
\newblock In \emph{Proceedings of the 2017 ACM on Conference on Information and
  Knowledge Management}, CIKM '17, pp.\  1747--1756, New York, NY, USA, 2017.
  ACM.
\newblock ISBN 978-1-4503-4918-5.
\newblock \doi{10.1145/3132847.3133010}.
\newblock URL \url{http://doi.acm.org/10.1145/3132847.3133010}.

\bibitem[Eric \& Manning(2017)Eric and Manning]{eric-manning:2017:EACLshort}
Mihail Eric and Christopher Manning.
\newblock A copy-augmented sequence-to-sequence architecture gives good
  performance on task-oriented dialogue.
\newblock In \emph{Proceedings of the 15th Conference of the European Chapter
  of the Association for Computational Linguistics: Volume 2, Short Papers},
  pp.\  468--473, Valencia, Spain, April 2017. Association for Computational
  Linguistics.
\newblock URL \url{http://www.aclweb.org/anthology/E17-2075}.

\bibitem[Eric et~al.(2017)Eric, Krishnan, Charette, and Manning]{ericKVR2017}
Mihail Eric, Lakshmi Krishnan, Francois Charette, and Christopher~D. Manning.
\newblock Key-value retrieval networks for task-oriented dialogue.
\newblock In \emph{Proceedings of the 18th Annual SIGdial Meeting on Discourse
  and Dialogue}, pp.\  37--49. Association for Computational Linguistics, 2017.
\newblock URL \url{http://aclweb.org/anthology/W17-5506}.

\bibitem[Graves et~al.(2014)Graves, Wayne, and Danihelka]{graves2014neural}
Alex Graves, Greg Wayne, and Ivo Danihelka.
\newblock Neural turing machines.
\newblock \emph{CoRR}, 2014.

\bibitem[Graves et~al.(2016)Graves, Wayne, Reynolds, Harley, Danihelka,
  Grabska-Barwi{\'n}ska, Colmenarejo, Grefenstette, Ramalho, Agapiou,
  et~al.]{graves2016hybrid}
Alex Graves, Greg Wayne, Malcolm Reynolds, Tim Harley, Ivo Danihelka, Agnieszka
  Grabska-Barwi{\'n}ska, Sergio~G{\'o}mez Colmenarejo, Edward Grefenstette,
  Tiago Ramalho, John Agapiou, et~al.
\newblock Hybrid computing using a neural network with dynamic external memory.
\newblock \emph{Nature}, 538\penalty0 (7626):\penalty0 471--476, 2016.

\bibitem[Gu et~al.(2016)Gu, Lu, Li, and Li]{guEtAl2016}
Jiatao Gu, Zhengdong Lu, Hang Li, and Victor~O.K. Li.
\newblock Incorporating copying mechanism in sequence-to-sequence learning.
\newblock In \emph{Proceedings of the 54th Annual Meeting of the Association
  for Computational Linguistics (Volume 1: Long Papers)}, pp.\  1631--1640,
  Berlin, Germany, August 2016. Association for Computational Linguistics.
\newblock URL \url{http://www.aclweb.org/anthology/P16-1154}.

\bibitem[Gulcehre et~al.(2016)Gulcehre, Ahn, Nallapati, Zhou, and
  Bengio]{gulcehreEtAl2016}
Caglar Gulcehre, Sungjin Ahn, Ramesh Nallapati, Bowen Zhou, and Yoshua Bengio.
\newblock Pointing the unknown words.
\newblock In \emph{Proceedings of the 54th Annual Meeting of the Association
  for Computational Linguistics (Volume 1: Long Papers)}, pp.\  140--149,
  Berlin, Germany, August 2016. Association for Computational Linguistics.
\newblock URL \url{http://www.aclweb.org/anthology/P16-1014}.

\bibitem[He et~al.(2017)He, Liu, Liu, and Zhao]{heEtAl2017Long1}
Shizhu He, Cao Liu, Kang Liu, and Jun Zhao.
\newblock Generating natural answers by incorporating copying and retrieving
  mechanisms in sequence-to-sequence learning.
\newblock In \emph{Proceedings of the 55th Annual Meeting of the Association
  for Computational Linguistics (Volume 1: Long Papers)}, pp.\  199--208,
  Vancouver, Canada, July 2017. Association for Computational Linguistics.
\newblock URL \url{http://aclweb.org/anthology/P17-1019}.

\bibitem[Kaiser et~al.(2017)Kaiser, Nachum, Roy, and Bengio]{KaiserNRB17}
Lukasz Kaiser, Ofir Nachum, Aurko Roy, and Samy Bengio.
\newblock Learning to remember rare events.
\newblock \emph{International Conference on Learning Representations}, 2017.

\bibitem[Kingma \& Ba(2015)Kingma and Ba]{KingmaB14}
Diederik~P Kingma and Jimmy Ba.
\newblock Adam: A method for stochastic optimization.
\newblock \emph{International Conference on Learning Representations}, 2015.

\bibitem[Lee \& Stent(2016)Lee and Stent]{lee2016task}
Sungjin Lee and Amanda Stent.
\newblock Task lineages: Dialog state tracking for flexible interaction.
\newblock In \emph{Proceedings of the 17th Annual Meeting of the Special
  Interest Group on Discourse and Dialogue}, pp.\  11--21, 2016.

\bibitem[Lei et~al.(2018)Lei, Jin, Kan, Ren, He, and Yin]{lei2018sequicity}
Wenqiang Lei, Xisen Jin, Min-Yen Kan, Zhaochun Ren, Xiangnan He, and Dawei Yin.
\newblock Sequicity: Simplifying task-oriented dialogue systems with single
  sequence-to-sequence architectures.
\newblock In \emph{Proceedings of the 56th Annual Meeting of the Association
  for Computational Linguistics (Volume 1: Long Papers)}, volume~1, pp.\
  1437--1447, 2018.

\bibitem[Liu \& Perez(2017)Liu and Perez]{perez2016gated}
Fei Liu and Julien Perez.
\newblock Gated end-to-end memory networks.
\newblock In \emph{Proceedings of the 15th Conference of the European Chapter
  of the Association for Computational Linguistics: Volume 1, Long Papers},
  pp.\  1--10, Valencia, Spain, April 2017. Association for Computational
  Linguistics.
\newblock URL \url{http://www.aclweb.org/anthology/E17-1001}.

\bibitem[Madotto et~al.(2018)Madotto, Wu, and Fung]{mem2seq}
Andrea Madotto, Chien-Sheng Wu, and Pascale Fung.
\newblock Mem2seq: Effectively incorporating knowledge bases into end-to-end
  task-oriented dialog systems.
\newblock In \emph{Proceedings of the 56th Annual Meeting of the Association
  for Computational Linguistics (Volume 1: Long Papers)}, pp.\  1468--1478.
  Association for Computational Linguistics, 2018.
\newblock URL \url{http://aclweb.org/anthology/P18-1136}.

\bibitem[Merity et~al.(2017)Merity, Xiong, Bradbury, and
  Socher]{merity2016pointer}
Stephen Merity, Caiming Xiong, James Bradbury, and Richard Socher.
\newblock Pointer sentinel mixture models.
\newblock \emph{International Conference on Learning Representations}, 2017.

\bibitem[Miller et~al.(2016)Miller, Fisch, Dodge, Karimi, Bordes, and
  Weston]{millerEtAl2016}
Alexander Miller, Adam Fisch, Jesse Dodge, Amir-Hossein Karimi, Antoine Bordes,
  and Jason Weston.
\newblock Key-value memory networks for directly reading documents.
\newblock In \emph{Proceedings of the 2016 Conference on Empirical Methods in
  Natural Language Processing}, pp.\  1400--1409, Austin, Texas, November 2016.
  Association for Computational Linguistics.
\newblock URL \url{https://aclweb.org/anthology/D16-1147}.

\bibitem[See et~al.(2017)See, Liu, and Manning]{seeliumanning2017Long}
Abigail See, Peter~J. Liu, and Christopher~D. Manning.
\newblock Get to the point: Summarization with pointer-generator networks.
\newblock In \emph{Proceedings of the 55th Annual Meeting of the Association
  for Computational Linguistics (Volume 1: Long Papers)}, pp.\  1073--1083,
  Vancouver, Canada, July 2017. Association for Computational Linguistics.
\newblock URL \url{http://aclweb.org/anthology/P17-1099}.

\bibitem[Seo et~al.(2017)Seo, Min, Farhadi, and Hajishirzi]{seo2016query}
Minjoon Seo, Sewon Min, Ali Farhadi, and Hannaneh Hajishirzi.
\newblock Query-reduction networks for question answering.
\newblock \emph{International Conference on Learning Representations}, 2017.

\bibitem[Serban et~al.(2016)Serban, Sordoni, Bengio, Courville, and
  Pineau]{serban2016building}
Iulian~Vlad Serban, Alessandro Sordoni, Yoshua Bengio, Aaron~C Courville, and
  Joelle Pineau.
\newblock Building end-to-end dialogue systems using generative hierarchical
  neural network models.
\newblock In \emph{AAAI}, pp.\  3776--3784, 2016.

\bibitem[Sharma et~al.(2016)Sharma, He, Suleman, Schulz, and
  Bachman]{sharma2016natural}
Shikhar Sharma, Jing He, Kaheer Suleman, Hannes Schulz, and Philip Bachman.
\newblock Natural language generation in dialogue using lexicalized and
  delexicalized data.
\newblock \emph{International Conference on Learning Representations}, 2016.

\bibitem[Su et~al.(2016)Su, Gasic, Mrksic, Rojas-Barahona, Ultes, Vandyke, Wen,
  and Young]{su2016line}
Pei-Hao Su, Milica Gasic, Nikola Mrksic, Lina Rojas-Barahona, Stefan Ultes,
  David Vandyke, Tsung-Hsien Wen, and Steve Young.
\newblock On-line active reward learning for policy optimisation in spoken
  dialogue systems.
\newblock \emph{Association for Computational Linguistics}, 2016.

\bibitem[Sukhbaatar et~al.(2015)Sukhbaatar, Weston, Fergus,
  et~al.]{sukhbaatar2015end}
Sainbayar Sukhbaatar, Jason Weston, Rob Fergus, et~al.
\newblock End-to-end memory networks.
\newblock In \emph{Advances in neural information processing systems}, pp.\
  2440--2448, 2015.

\bibitem[Vaswani et~al.(2017)Vaswani, Shazeer, Parmar, Uszkoreit, Jones, Gomez,
  Kaiser, and Polosukhin]{vaswani2017attention}
Ashish Vaswani, Noam Shazeer, Niki Parmar, Jakob Uszkoreit, Llion Jones,
  Aidan~N Gomez, {\L}ukasz Kaiser, and Illia Polosukhin.
\newblock Attention is all you need.
\newblock In \emph{Advances in Neural Information Processing Systems}, pp.\
  6000--6010, 2017.

\bibitem[Vinyals et~al.(2015)Vinyals, Fortunato, and
  Jaitly]{vinyals2015pointer}
Oriol Vinyals, Meire Fortunato, and Navdeep Jaitly.
\newblock Pointer networks.
\newblock In C.~Cortes, N.~D. Lawrence, D.~D. Lee, M.~Sugiyama, and R.~Garnett
  (eds.), \emph{Advances in Neural Information Processing Systems 28}, pp.\
  2692--2700. Curran Associates, Inc., 2015.
\newblock URL \url{http://papers.nips.cc/paper/5866-pointer-networks.pdf}.

\bibitem[Wang et~al.(2016)Wang, Lu, Li, and Liu]{wangEtAl2016}
Mingxuan Wang, Zhengdong Lu, Hang Li, and Qun Liu.
\newblock Memory-enhanced decoder for neural machine translation.
\newblock In \emph{Proceedings of the 2016 Conference on Empirical Methods in
  Natural Language Processing}, pp.\  278--286, Austin, Texas, November 2016.
  Association for Computational Linguistics.
\newblock URL \url{https://aclweb.org/anthology/D16-1027}.

\bibitem[Wen et~al.(2017)Wen, Gasic, Mrksic, Rojas-Barahona, hao Su, Ultes,
  Vandyke, and Young]{wen2016network}
Tsung-Hsien Wen, Milica Gasic, Nikola Mrksic, Lina~Maria Rojas-Barahona, Pei
  hao Su, Stefan Ultes, David Vandyke, and Steve~J. Young.
\newblock A network-based end-to-end trainable task-oriented dialogue system.
\newblock In \emph{EACL}, 2017.

\bibitem[Williams \& Young(2007)Williams and Young]{williams2007partially}
Jason~D Williams and Steve Young.
\newblock Partially observable markov decision processes for spoken dialog
  systems.
\newblock \emph{Computer Speech \& Language}, 21\penalty0 (2):\penalty0
  393--422, 2007.

\bibitem[Wu et~al.(2017)Wu, Madotto, Winata, and Fung]{wu2017dstc6}
Chien-Sheng Wu, Andrea Madotto, Genta Winata, and Pascale Fung.
\newblock End-to-end recurrent entity network for entity-value independent
  goal-oriented dialog learning.
\newblock In \emph{Dialog System Technology Challenges Workshop, DSTC6}, 2017.

\bibitem[Wu et~al.(2018)Wu, Madotto, Winata, and Fung]{dqmem8461426}
Chien-Sheng Wu, Andrea Madotto, Genta Winata, and Pascale Fung.
\newblock End-to-end dynamic query memory network for entity-value independent
  task-oriented dialog.
\newblock In \emph{2018 IEEE International Conference on Acoustics, Speech and
  Signal Processing (ICASSP)}, pp.\  6154--6158, April 2018.

\bibitem[Young et~al.(2013)Young, Ga{\v{s}}i{\'c}, Thomson, and
  Williams]{young2013pomdp}
Steve Young, Milica Ga{\v{s}}i{\'c}, Blaise Thomson, and Jason~D Williams.
\newblock Pomdp-based statistical spoken dialog systems: A review.
\newblock \emph{Proceedings of the IEEE}, 101\penalty0 (5):\penalty0
  1160--1179, 2013.

\bibitem[Zhao et~al.(2017)Zhao, Lu, Lee, and Eskenazi]{zhao2017generative}
Tiancheng Zhao, Allen Lu, Kyusong Lee, and Maxine Eskenazi.
\newblock Generative encoder-decoder models for task-oriented spoken dialog
  systems with chatting capability.
\newblock In \emph{Proceedings of the 18th Annual SIGdial Meeting on Discourse
  and Dialogue}, pp.\  27--36. Association for Computational Linguistics,
  August 2017.
\newblock URL \url{http://aclweb.org/anthology/W17-5505}.

\bibitem[Zhong et~al.(2018)Zhong, Xiong, and Socher]{zhong2018global}
Victor Zhong, Caiming Xiong, and Richard Socher.
\newblock Global-locally self-attentive dialogue state tracker.
\newblock In \emph{Association for Computational Linguistics}, 2018.

\end{thebibliography}
\bibliographystyle{iclr2019_conference}

\newpage
\appendix
\section{Tables}

\subsection{Training parameters}

\begin{table}[h]
\centering
\setlength{\belowcaptionskip}{5pt} 
\caption{Selected hyper-parameters in each dataset for different hops. The values is the embedding dimension and the GRU hidden size, and the values between parenthesis is the dropout rate. For all the models we used learning rate equal to 0.001, with a decay rate of 0.5. }
\begin{tabular}{|r|c|c|c|c|c|c|}
\hline
\multicolumn{1}{|l|}{} & T1 & T2 & T3 & T4 & T5 & SMD \\ \hline
GLMP K1 & 64 (0.1) & 64 (0.3) & 64 (0.3) & 64 (0.7) & 128 (0.3) & 128 (0.2) \\ \hline
GLMP K3 & 64 (0.3) & 64 (0.3) & 64 (0.3) & 64 (0.7) & 128 (0.1) & 128 (0.2) \\ \hline
GLMP K6 & 64 (0.3) & 64 (0.3) & 64 (0.5) & 64 (0.5) & 128 (0.1) & 128 (0.3) \\ \hline
\end{tabular}
\end{table}

\subsection{Dataset Statistics}
\begin{table}[h]
\setlength{\belowcaptionskip}{5pt} 
\caption{Dataset statistics for 2 datasets.}
\centering
\resizebox{\linewidth}{!}{
\begin{tabular}{|r|c|c|c|c|c|c|c|c|}
\hline
\textbf{Task} & \textbf{1} & \textbf{2} & \textbf{3} & \textbf{4} & \textbf{5} & \multicolumn{3}{c|}{\textbf{SMD}} \\ \cline{7-9} 
\multicolumn{1}{|l|}{} & \multicolumn{1}{l|}{} & \multicolumn{1}{l|}{} & \multicolumn{1}{l|}{} & \multicolumn{1}{l|}{} & \multicolumn{1}{l|}{} & \multicolumn{1}{l|}{Calendar} & \multicolumn{1}{l|}{Weather} & \multicolumn{1}{l|}{Navigation} \\ \hline
\textit{Avg. User turns} & 4 & 6.5 & 6.4 & 3.5 & 12.9 & \multicolumn{3}{c|}{2.6} \\ \hline
\textit{Avg. Sys turns} & 6 & 9.5 & 9.9 & 3.5 & 18.4 & \multicolumn{3}{c|}{2.6} \\ \hline
\textit{Avg. KB results} & 0 & 0 & 24 & 7 & 23.7 & \multicolumn{3}{c|}{66.1} \\ \hline
\textit{Avg. Sys words} & 6.3 & 6.2 & 7.2 & 5.7 & 6.5 & \multicolumn{3}{c|}{8.6} \\ \hline
\textit{Max. Sys words} & 9 & 9 & 9 & 8 & 9 & \multicolumn{3}{c|}{87} \\ \hline
\textit{Nb. Slot Types} & \multicolumn{5}{c|}{7} & 6 & 4 & 5 \\ \hline
\textit{Nb. Distinct Slot values} & \multicolumn{5}{c|}{-} & 79 & 65 & 140 \\ \hline
\textit{Vocabulary} & \multicolumn{5}{c|}{3747} & \multicolumn{3}{c|}{1601} \\ \hline
\textit{Train dialogues} & \multicolumn{5}{c|}{1000} & \multicolumn{3}{c|}{2425} \\ \hline
\textit{Val dialogues} & \multicolumn{5}{c|}{1000} & \multicolumn{3}{c|}{302} \\ \hline
\textit{Test dialogues} & \multicolumn{5}{c|}{1000 + 1000 OOV} & \multicolumn{3}{c|}{304} \\ \hline
\textit{Total Nb. Dialogues} & 4000 & 4000 & 4000 & 4000 & 4000 & 1034 & 997 & 1000 \\ \hline
\end{tabular}
}
\label{Tab:statistic}
\end{table}

\subsection{Human Evaluation}
\begin{figure}[h]
\centering
\includegraphics[width=0.6\linewidth]{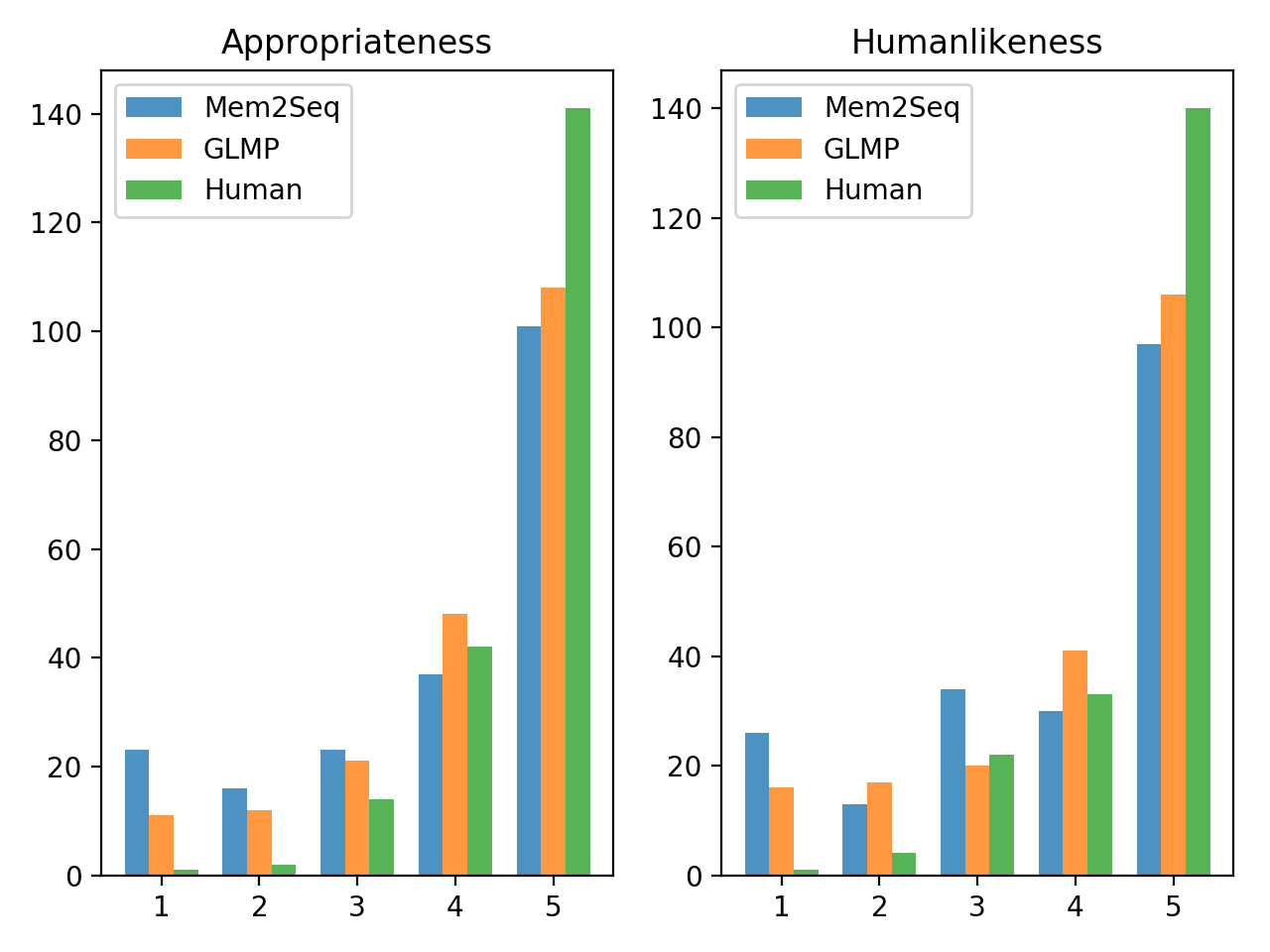} 
\caption{Appropriateness and human-likeness scores according to 200 dialogue scenarios. }
\end{figure}

Appropriateness \\
5: Correct grammar, correct logic, correct dialogue flow, and correct entity provided \\
4: Correct dialogue flow, logic and grammar but has slightly mistakes in entity provided \\
3: Noticeable mistakes about grammar or logic or entity provided but acceptable \\
2: Poor grammar, logic and entity provided \\
1: Wrong grammar, wrong logic, wrong dialogue flow, and wrong entity provided

Human-Likeness (Naturalness) \\
5: The utterance is 100\% like what a person will say \\
4: The utterance is 75\% like what a person will say \\
3: The utterance is 50\% like what a person will say \\
2: The utterance is 25\% like what a person will say \\
1:  The utterance is 0\% like what a person will say \\

\section{Error Analysis}
For bAbI dialogues, the mistakes are mainly from task 3, which is recommending restaurants based on their rating from high to low. We found that sometimes the system will keep sending those restaurants with the higher score even if the user rejected them in the previous turns. On the other hand, SMD is more challenging for response generation. First, we found that the model makes mistakes when the KB has several options corresponding to the user intention. For example, once the user has more than one doctor appointment in the table, the model can barely recognize. In addition, since we do not include the domain specific and user intention supervision, wrong delexicalized responses may be generated, which results in an incorrect entity copy. Lastly, we found that the copied entities may not be matched to the generated sketch tags. For example, an address tag may result in a distance entity copy. We leave the space of improvement to future works.

\section{Additional Discussion}
One of the reviewers suggested us to compare our work to some existing dialogue framework such as PyDial~\footnote{http://www.camdial.org/pydial/}. To the best of our knowledge, in the PyDial framework, it requires to have the dialogue act’s labels for the NLU module and the belief states’ labels for the belief tracker module. The biggest challenge is we do not have such labels in the SMD and bAbI datasets. Moreover, the semi tracker in PyDial is rule-based, which need to re-write rules whenever it encounters a new domain or new datasets. Even its dialogue management module could be a learning solution like policy networks, the input of the policy network is still the hand-crafted state features and labels. Therefore, without the rules and labels predefined in the NLU and belief tracker modules, PyDial could not learn a good policy network. 

Truly speaking, based on the data we have (not very big size) and the current state-of-the-art machine learning algorithms and models, we believe that a well and carefully constructed task-oriented dialogue system using PyDial in a known domain using human rules (in NLU and Belief Tracker) with policy networks may outperform the end-to-end systems (more robust). However, in this paper, without additional human labels and human rules, we want to explore the potential and the advantage of end-to-end systems. Besides easy to train, for multi-domain cases, or even zero-shot domain cases, we believe end-to-end approaches will have better adaptability compared to any rule-based systems.

\pagebreak

\section{Visualization}
\begin{figure}[h]
\centering
\includegraphics[width=0.65\linewidth]{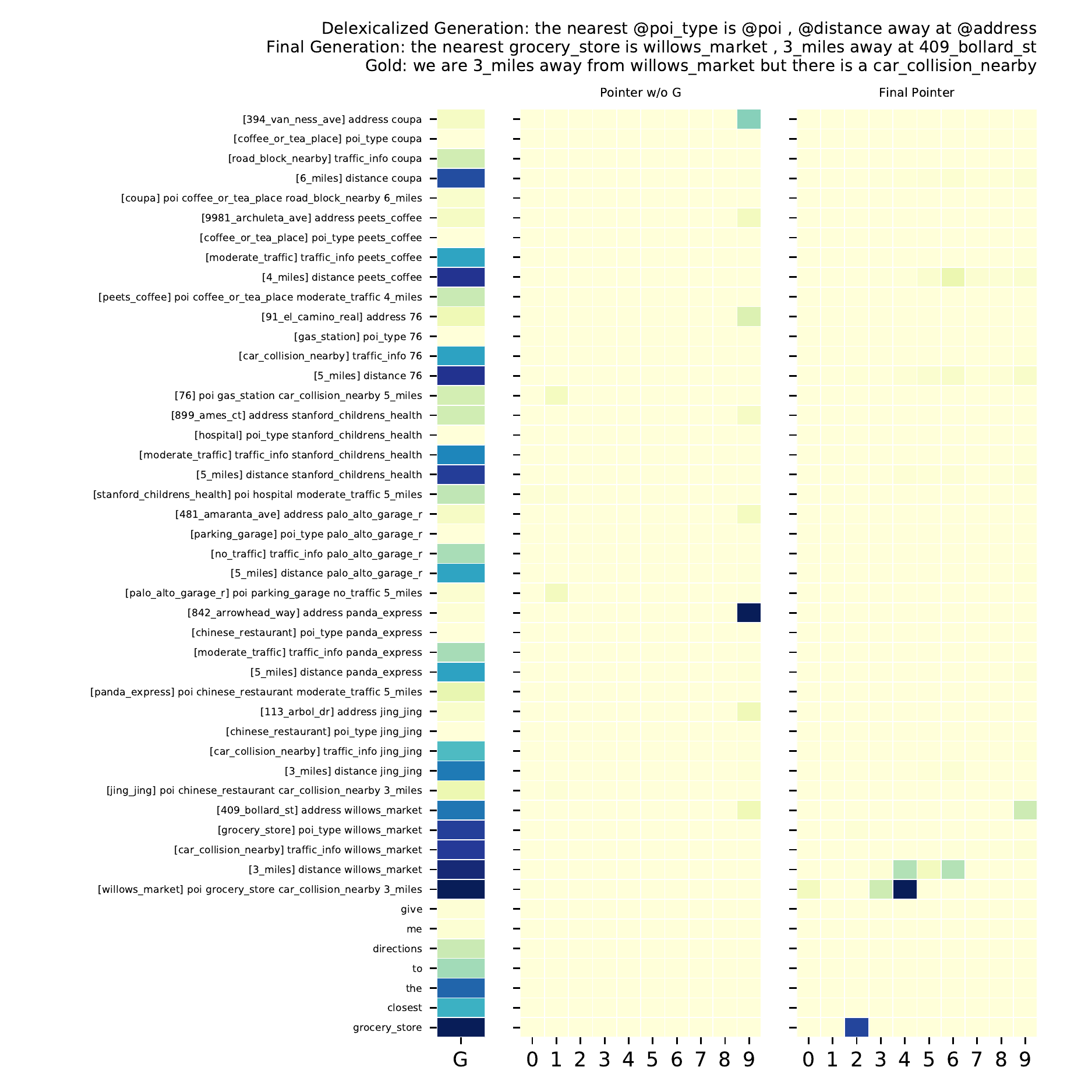} 
\caption{Memory attention visualization from the SMD navigation domain.}
\end{figure}

\begin{figure}[h]
\centering
\includegraphics[width=0.65\linewidth]{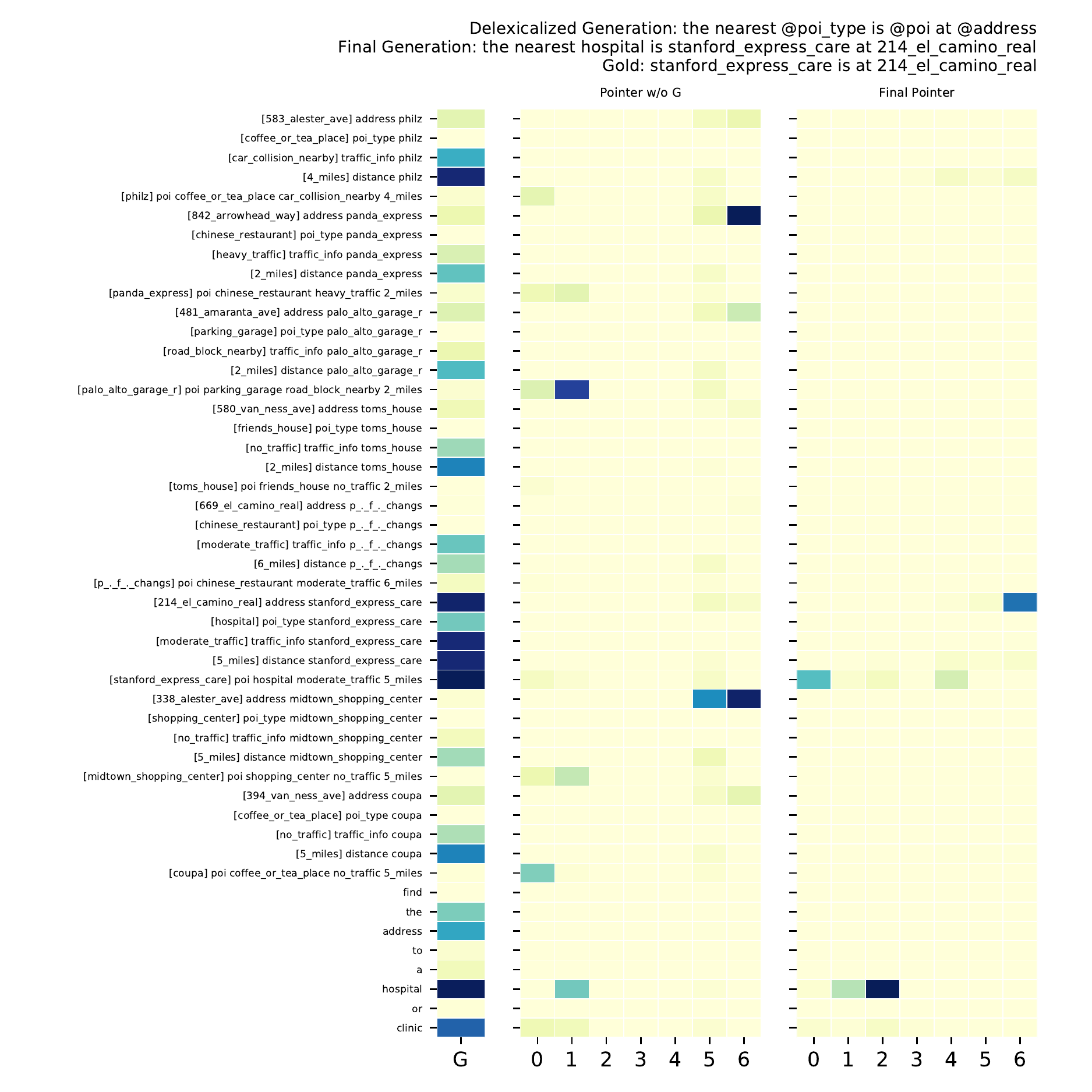} 
\caption{Memory attention visualization from the SMD navigation domain.}
\end{figure}

\begin{figure}[h]
\centering
\includegraphics[width=0.65\linewidth]{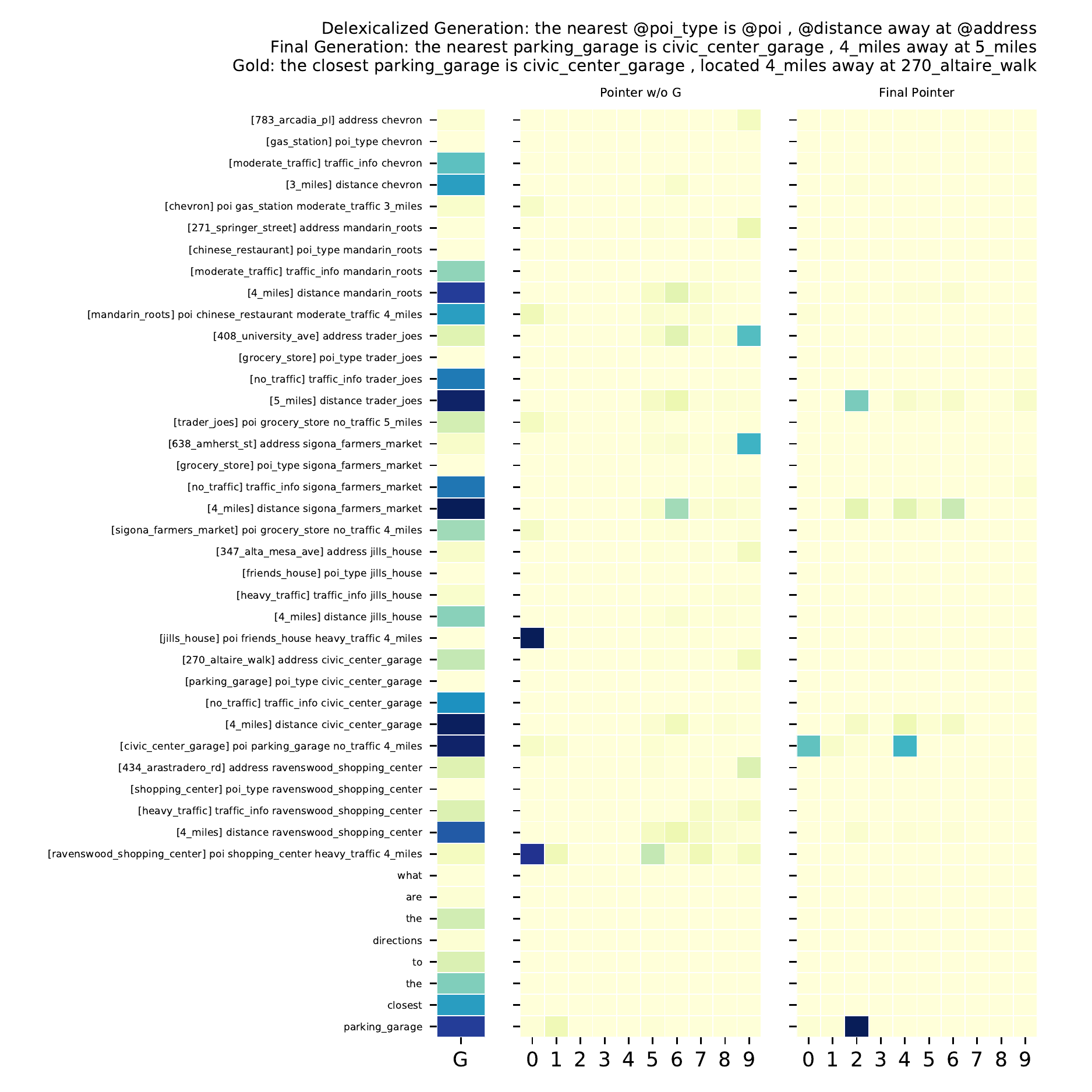} 
\caption{Memory attention visualization from the SMD navigation domain.}
\end{figure}

\begin{figure}[h]
\centering
\includegraphics[width=0.65\linewidth]{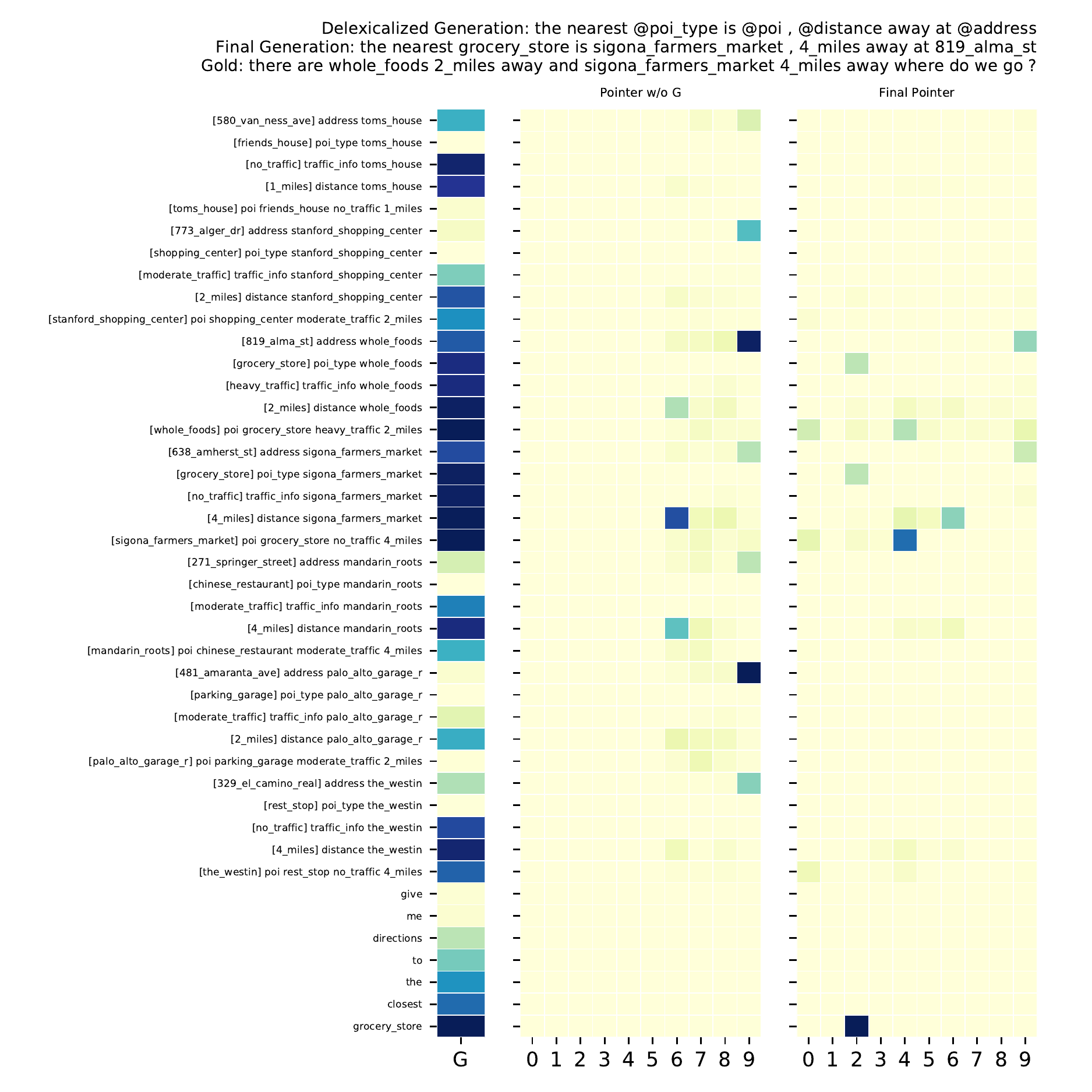} 
\caption{Memory attention visualization from the SMD navigation domain.}
\end{figure}

\begin{figure}[h]
\centering
\includegraphics[width=0.65\linewidth]{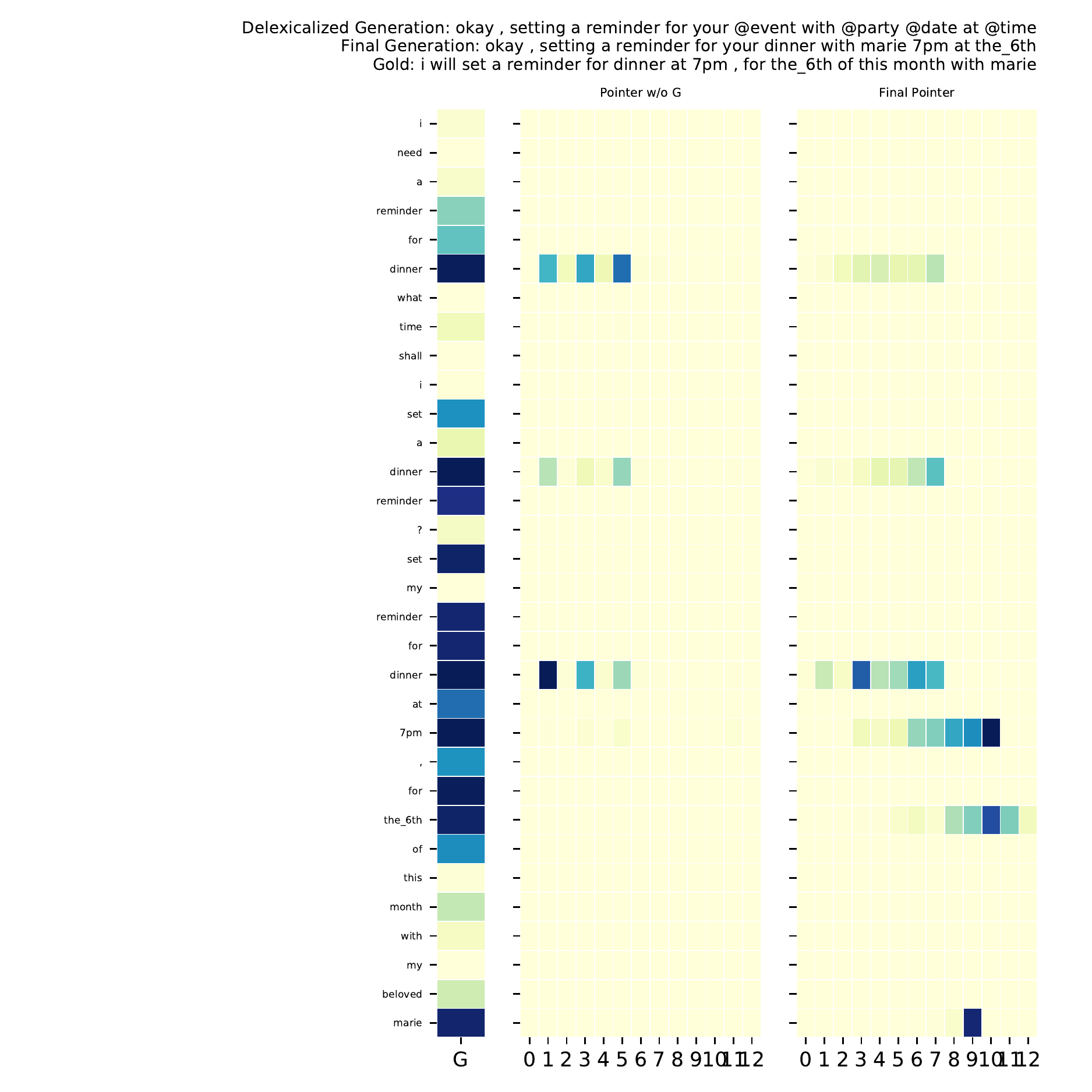} 
\caption{Memory attention visualization from the SMD schedule domain.}
\end{figure}

\begin{figure}[h]
\centering
\includegraphics[width=0.65\linewidth]{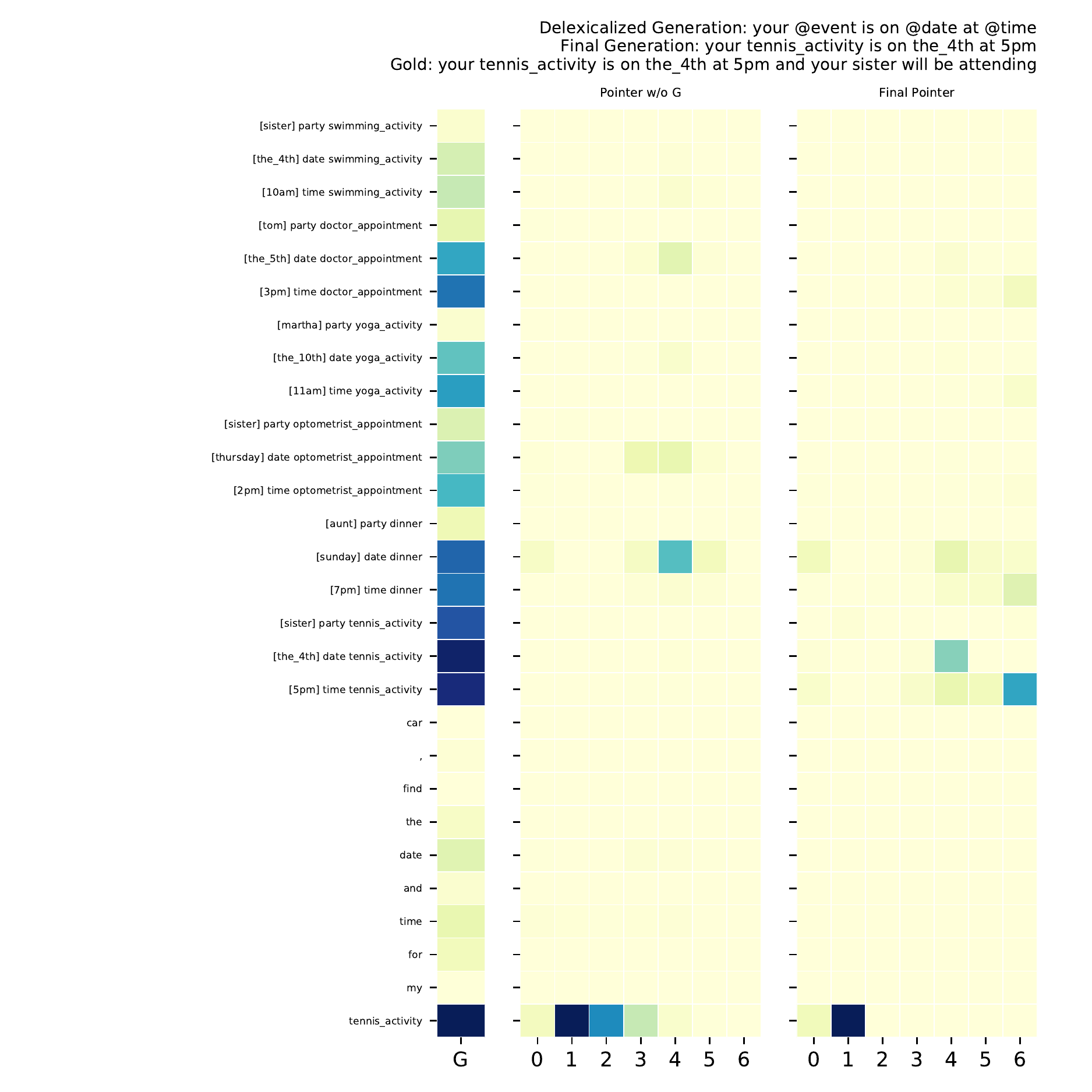} 
\caption{Memory attention visualization from the SMD schedule domain.}
\end{figure}

\begin{figure}[h]
\centering
\includegraphics[width=0.65\linewidth]{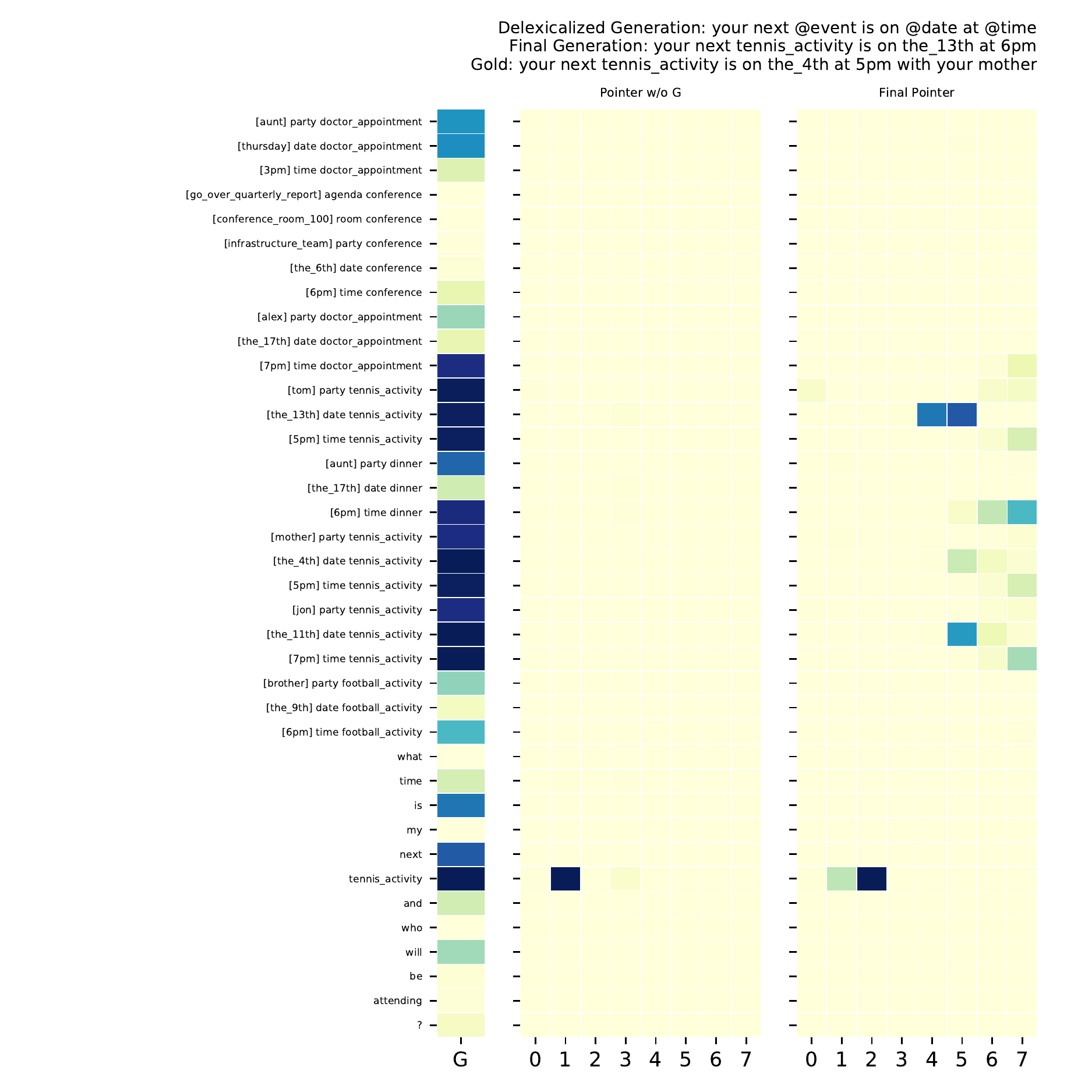} 
\caption{Memory attention visualization from the SMD schedule domain.}
\end{figure}

\begin{figure}[h]
\centering
\includegraphics[width=0.65\linewidth]{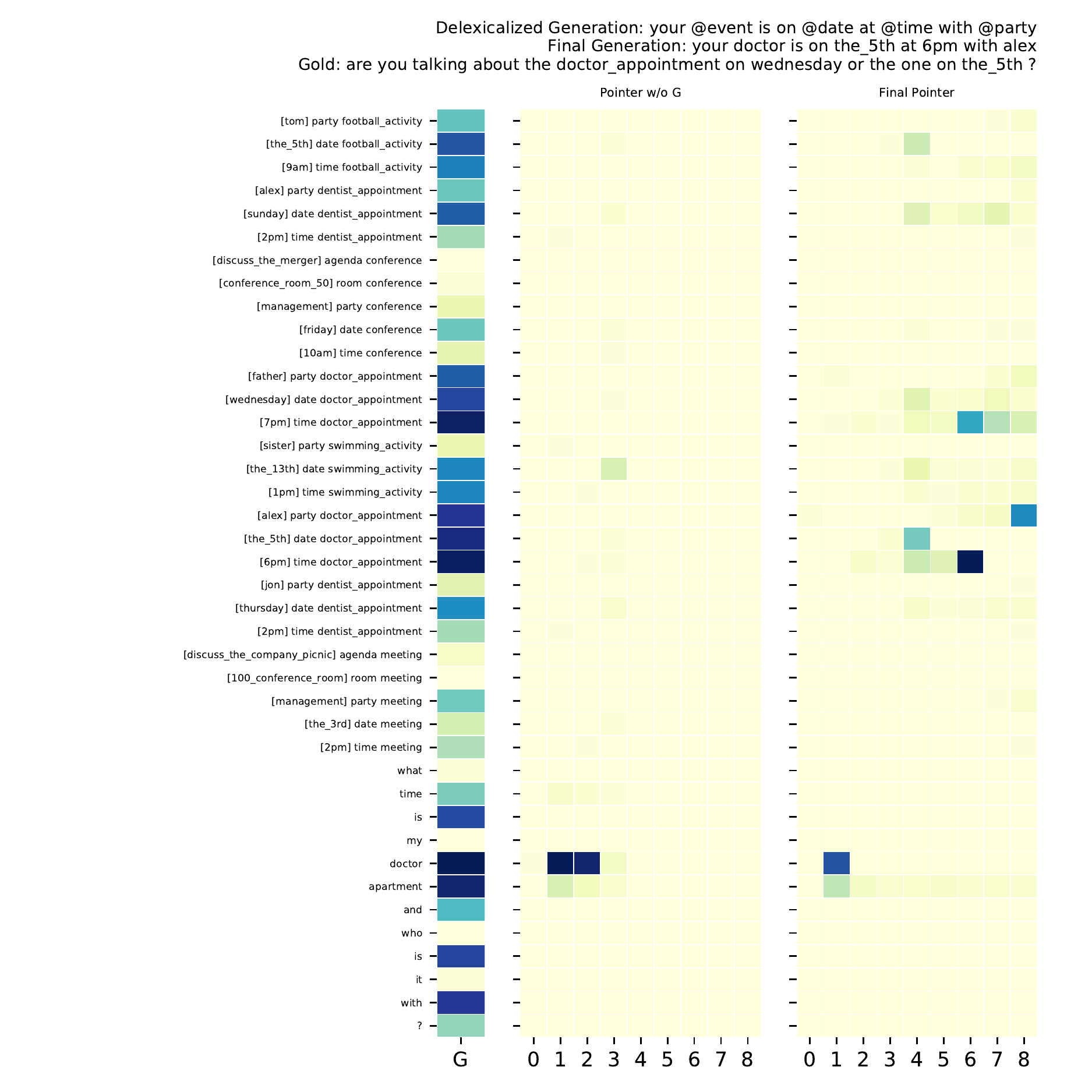} 
\caption{Memory attention visualization from the SMD schedule domain.}
\end{figure}

\begin{figure}[h]
\centering
\includegraphics[height=\textheight]{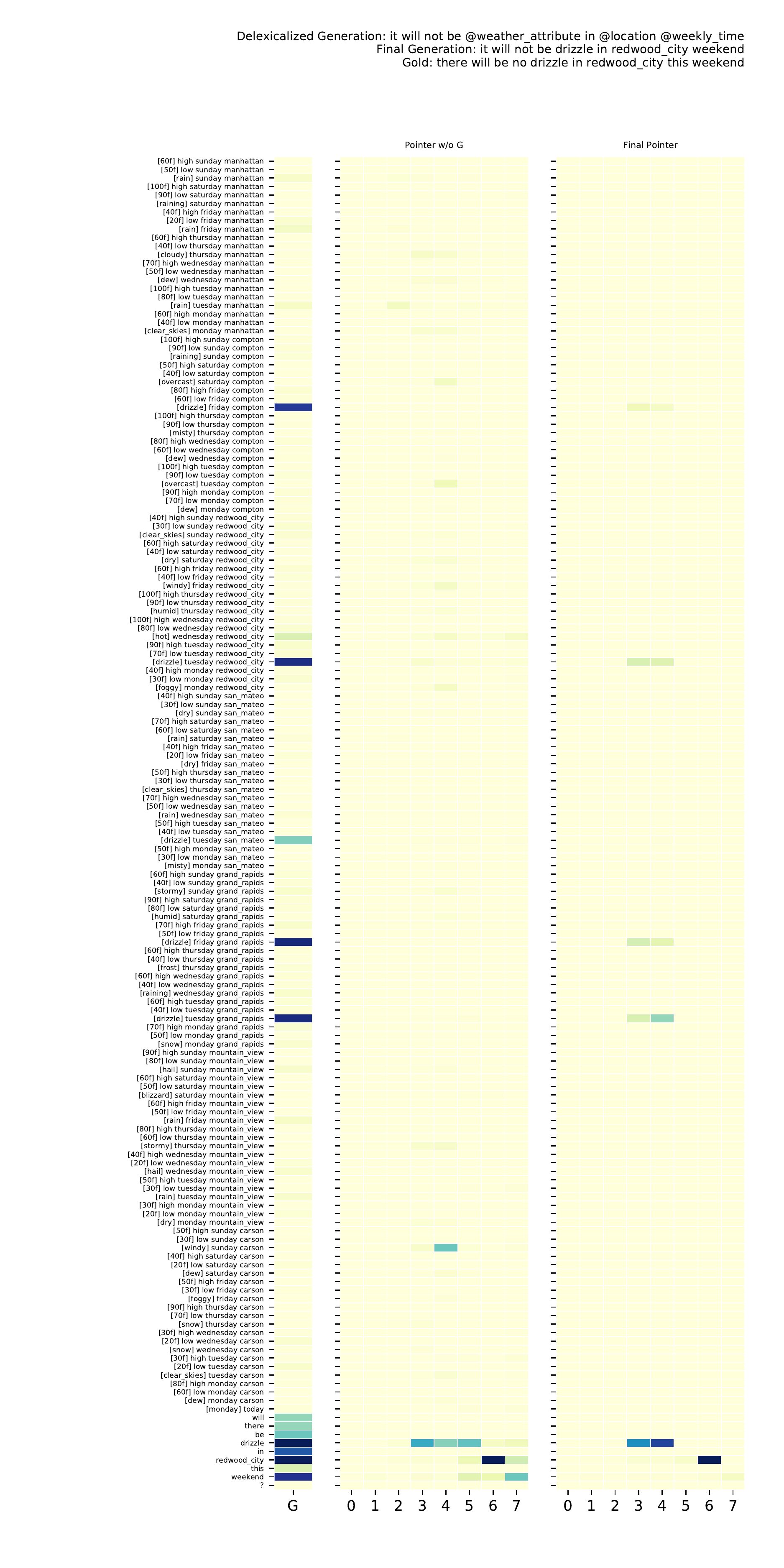} 
\caption{Memory attention visualization from the SMD weather domain.}
\setlength{\abovecaptionskip}{-5pt} 
\end{figure}

\begin{figure}[h]
\centering
\includegraphics[height=\textheight]{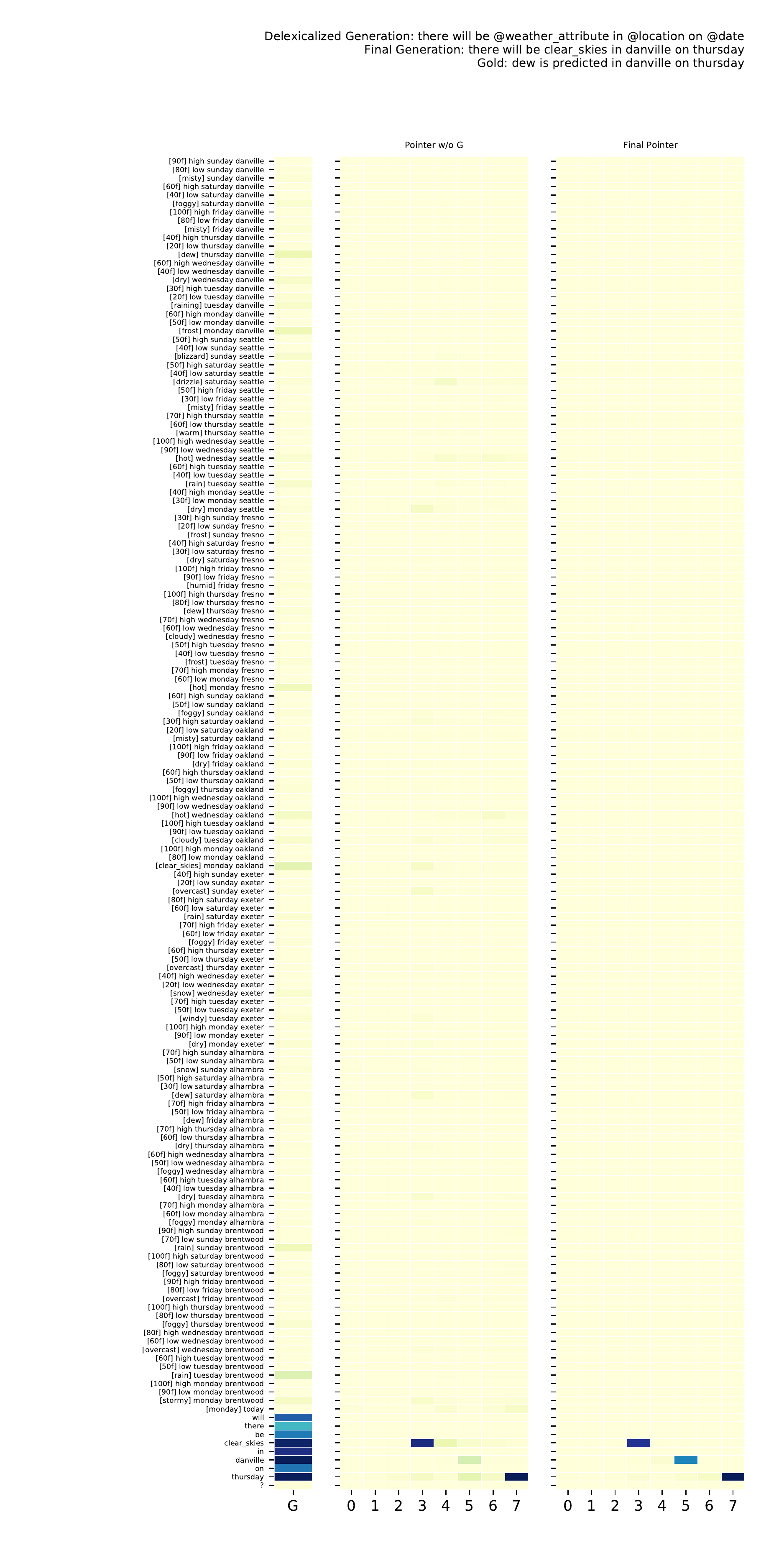} 
\caption{Memory attention visualization from the SMD weather domain.}
\setlength{\abovecaptionskip}{-5pt} 
\end{figure}

\end{document}